\pdfoutput=1
\documentclass[11pt]{article}
\usepackage[final]{acl}

\usepackage{times}
\usepackage{latexsym}
\usepackage{amsmath}

\usepackage[T1]{fontenc}
\usepackage[utf8]{inputenc}

\usepackage{microtype}

\usepackage{inconsolata}

\usepackage{graphicx}
\usepackage{subcaption} 

\usepackage{tikz}
\usetikzlibrary{positioning,decorations.pathreplacing}
\usetikzlibrary{trees, shapes, arrows.meta, positioning}

\usepackage[utf8]{inputenc}
\usepackage{forest}
\usepackage{geometry}
\usepackage{forest}

\usepackage{colortbl}
\usepackage{pifont}
\definecolor{LightRed}{RGB}{255,200,200}

\title{An Empirical Survey of Model Merging Algorithms \\ for Social Bias Mitigation}

\author{
    Daiki Shirafuji$^{1}$, Tatsuhiko Saito$^{1}$, Yasutomo Kimura$^{2}$ \\
    $^{1}$ Mitsubishi Electric Corporation \\
    $^{2}$ Otaru University of Commerce \\
    \texttt{\{Shirafuji.Daiki@ay, Saito.Tatsuhiko@db\}.MitsubishiElectric.co.jp},\\
    \texttt{kimura@res.otaru-uc.ac.jp} \\
}

\begin{document}

\maketitle

\begin{abstract}

Large language models (LLMs) are known to inherit and even amplify societal biases present in their pre-training corpora, threatening fairness and social trust. To address this issue, recent work has explored ``editing'' LLM parameters to mitigate social bias with model merging approaches; however, there is no empirical comparison. In this work, we empirically survey seven algorithms: Linear, Karcher Mean, SLERP, NuSLERP, TIES, DELLA, and Nearswap, applying 13 open weight models in the GPT, LLaMA, and Qwen families. We perform a comprehensive evaluation using three bias datasets (BBQ, BOLD, and HONEST) and measure the impact of these techniques on LLM performance in downstream tasks of the SuperGLUE benchmark. We find a trade-off between bias reduction and downstream performance: methods achieving greater bias mitigation degrade accuracy, particularly on tasks requiring reading comprehension and commonsense and causal reasoning. Among the merging algorithms, Linear, SLERP, and Nearswap consistently reduce bias while maintaining overall performance, with SLERP at moderate interpolation weights emerging as the most balanced choice. These results highlight the potential of model merging algorithms for bias mitigation, while indicating that excessive debiasing or inappropriate merging methods may lead to the degradation of important linguistic abilities.

\end{abstract}

\textcolor{red}{\textit{Warning: This paper contains examples that may be considered discriminatory}.}

\section{Introduction}
Large language models (LLMs) have recently achieved remarkable performance in various tasks in natural language processing~\citep{gpt4,qwen25}.
However, some studies~\citep{Bolukbasi,Navigli,coli_a_00524} have pointed out that
social biases \footnote{\citet{Navigli} define \textit{ biases} in the field of natural language processing as
``prejudices, stereotypes, and discriminatory attitudes against certain groups of people,'' and we also adopt this definition throughout this paper.} embedded in pre-training data are often mirrored in model outputs.
These works have shown that LLMs exhibit negative biases toward various social attributes, such as gender, race, or religion.
Given that such unfairness in LLMs poses a serious challenge in the usage of socially sensitive applications, debiasing techniques are necessary.

Previous work on reducing social bias has explored various approaches, such as training LLMs with synthetic examples \citep{zmigrod-etal-2019-counterfactual,ravfogel-etal-2020-null,Schick_self_debias}.
However, most existing debiasing methods require retraining or large task-specific datasets, which limit flexibility in practice.

For this reason, model merging \citep{Model_soups}, which fuses multiple fine-tuned checkpoints originating from the same initialization directly in parameter space, has recently been explored to mitigate social bias,
such as methods based on simple task arithmetic \citep{shirafuji-etal-2025-bias} or parameter selective editing \citep{lutz-etal-2024-local}.

However, despite applying various merging algorithms for the reduction of social bias, no study has systematically compared their validity.

\begin{figure}[t]
  \centering
    \includegraphics[width=\columnwidth]{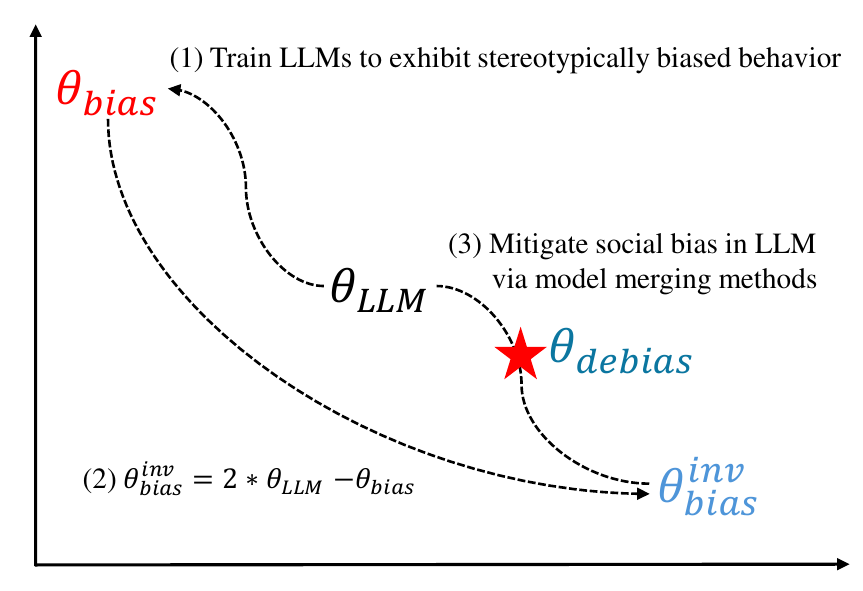}
    \caption{An overview of social bias mitigation process based on model merging methods.}
    \label{fig:overview}
\end{figure}

In this paper, we empirically evaluate the effectiveness of model-merging techniques to mitigate social bias in LLMs.
An overview of our pipeline is illustrated in Figure~\ref{fig:overview}.
According to \citet{shirafuji-etal-2025-bias}, we first fine-tune a pre-trained LLM on biased data,
thereby amplifying social bias in the model, and extract the difference in parameters between the pre-trained LLM and the biased LLM as the bias vector.
Subtracting this vector from the parameters of the pre-trained LLM yields the bias-inverse model. We then merge with the original pre-trained model and the inverse model using various algorithms.

Empirical experiments are conducted for seven merging techniques:
Linear \citep{Model_soups}, Karcher Mean \citep{grove1973conjugate},
SLERP \citep{slerp}, NuSLERP \citep{mergekit},
TIES \citep{TIES}, DELLA \citep{deep2024dellamerging},
%breadcrumbs \citep{breadcrumbs},
and Nearswap \citep{mergekit}.
We evaluated 13 models that are in the GPT \citep{gpt2,gptneo}, LLaMA \citep{llama2,llama3}, and Qwen \citep{qwen2} families.
Performances are measured in three bias datasets (BBQ \citep{bbq}, BOLD \citep{bold}, and HONEST \citep{honest})
and, to ensure downstream quality is preserved, on the SuperGLUE benchmark \citep{SuperGLUE}.

Our contributions are as follows:
\begin{itemize}
    \item Conducting an empirical survey on seven model merging algorithms for social bias mitigation with three bias benchmarks and SuperGLUE across 13 LLMs.
    \item Identifying SLERP with moderate interpolation weights as the most balanced method, achieving effective bias reduction without sacrificing downstream accuracy.
    \item Highlighting the necessity of verifying performance on tasks such as reading comprehension and commonsense / causal reasoning for social bias mitigation.
\end{itemize}

\section{Related Works}

\subsection{Model Merging Algorithms}
Recently,
model merging has emerged as an effective strategy for combining the strengths of multiple models without expensive retraining
\citep{li2023deep,yang2024model}.
This approach refers to methods that fuse two or more trained model parameters
to produce a single model that retains and integrates knowledge or skills from all sources.

Model merging is pioneered by the linear averaging method (``linear''),
treating weights as vectors
and simply merged by arithmetic means \citep{Model_soups}.
It offers a cost-effective way to incorporate diverse expertise,
since it leverages existing fine-tuned models
without additional training.
Some studies \citep{matena2022merging,lee-etal-2025-dynamic} generalize this idea
by weighting each parameter inversely to its Fisher information,
resulting in combinations consistent with likelihood.

Merging methods based on sphere interpolation
\citep{slerp,mergekit,grove1973conjugate}
regard parameter vectors as lying on a sphere. SLERP \citep{slerp} performs an interpolation between two models, and the Karcher Mean \citep{grove1973conjugate} iteratively finds the Riemannian centroid for any number of models.
NuSLERP \citep{mergekit} adds per-tensor normalization to correct for norm drift.

Inspired by these model merging approaches,
\citet{ilharco2022editing} proposed the task arithmetic approach under the concept of ``task vector.''
Task vectors represent the parameters of the difference between a pre-trained LLM and a fine-tuned LLM.

TIES-Merging \citep{TIES} resets tiny deltas, resolves sign conflicts, and then linearly combines cleaned updates;
DELLA-Merging \citep{deep2024dellamerging} is also a model merging technique that orders parameters by magnitude, preferentially removes smaller ones, and rescales the remaining values to balance the model.

\subsection{Model Merging for Social Bias Mitigation}
Some studies have demonstrated that merging algorithms can substantially reduce social bias while preserving performance in downstream tasks.
\citet{shirafuji-etal-2025-bias} construct a bias vector from the bias-amplifying corpora, subtract it from the base model, and extract bias parameters.
\citet{dige-etal-2024-machine} show that simply negating a task vector trained on biased data rivals heavier unlearning objectives for LLaMA-2.
\citet{gao-etal-2024-ethos} refine this idea by projecting the raw vector onto an orthogonal subspace before subtraction, thus preserving general linguistic skills.

Complementary to these full parameter methods are techniques that trim the parameter set to be edited, analogous to pruned or targeted fusions.  
\citet{lutz-etal-2024-local} locate fewer than 0.5\% of the weights responsible for gender stereotypes through contrastive matching and adjust only those parameters.  
LoRA‑based subtraction \citep{ki-etal-2024-inspecting} and the two-stage selective knowledge unlearning of \citet{liu-etal-2024-towards-safer} follow a similar philosophy: first isolate harmful knowledge in a compact adapter, then merge or subtract it from the backbone.  
Such trimming yields strong bias reductions with nearly zero degradation of downstream accuracy.

A third line of work takes advantage of mechanistic insights to pinpoint bias‑bearing components before editing.  
Neuronal interventions at the neuron level of \citet{garnier2024decompound} disable gender-sensitive circuits by setting their activations to zero, while \citet{qin2025lftf} calculate the bias contribution of each transformer block and fine-tune only the most culpable layer.
These interpretable edits modify the parameters \(\ll 1\%\) yet mitigate social bias in the Winogender \citep{rudinger-etal-2018-gender} and StereoSet \citep{nadeem-etal-2021-stereoset} datasets, confirming that social biases are often concentrated in identifiable substructures.

\paragraph{Impact on Downstream Tasks.}
Across all categories, careful parameter merges incur little collateral damage:
\citet{shirafuji-etal-2025-bias} report a 3\% drop on average in GLUE benchmarks \citep{wang2018glue}, but they also observe over 50\% declines in the COLA dataset. 
\citet{dige-etal-2024-machine} find no significant increase in perplexity
and both \citet{lutz-etal-2024-local} and \citet{gao-etal-2024-ethos} observe unchanged or even improved accuracy in the downstream tasks.
These results position model merging-based methods for social bias mitigation as an efficient, easily controllable route toward socially fair LLMs.

Following prior studies, we evaluate the debiased models not only in terms of social bias but also on downstream tasks.
Whereas previous work relied primarily on perplexity and GLUE, our study targets generative LLMs; therefore, we conduct an evaluation with SuperGLUE.

\section{Merging Experiments for Debiasing}

\subsection{Preliminary Preparations for Model Merging} \label{3-1}
In this section, we describe the preparations for applying model merging to mitigate social bias.

Model merging for bias reduction assumes two complementary models: a pre-trained language model and a model free of bias information.
However, presuming the availability of such a pre-debiased model is a flawed premise.

Therefore, in this study, we adopt the approach of \citet{shirafuji-etal-2025-bias}, which inverts the information of bias within the LLM using task arithmetic \citep{ilharco2022editing}.  
The overview of this process is shown in Figure~\ref{fig:overview}.
Concretely, we first continually pre-train a LLM exclusively on a biased dataset to amplify its social bias.
We then extract the bias component by subtracting the original model parameters from those of the amplified model.
Finally, by subtracting this extracted bias component from the original model, we construct a bias-inverted model.
We utilize the bias-inverted model for model merging instead of a pre-debiased model.

In detail, this process is expressed by the following equation.

\begin{equation}
    \begin{aligned}
      \theta_{bias}^{inv}
      &= \theta_{LLM} - \theta_{BV} \\
      &= \theta_{LLM} - (\theta_{bias} - \theta_{LLM}) \\
      &= 2 \, \theta_{LLM} - \theta_{bias},
    \end{aligned}
\end{equation}
where $\theta_{LLM}$, $\theta_{bias}$, $\theta_{BV}$, and $\theta_{bias}^{inv}$
are the parameters of pre-trained LLMs, bias-amplified models, social bias components, and bias-inverted models, respectively.

\subsection{Model Merging for Debiasing}
\subsubsection{Merging Formulation}
In this section, we describe the way to construct debiased LLMs based on model merging approaches.

The formula of debiasing is described below:
\begin{equation}
    \begin{aligned}
      \theta_{debias} = (1-\alpha) \, \theta_{LLM} \, \textcircled{+} \, \alpha \, \theta_{bias}^{inv},
    \end{aligned}
    \label{eq:debias}
\end{equation}
where $\theta_{debias}$ represents the debiased LLM parameter,
and $\alpha$ denotes the scaling weight of $\theta_{bias}^{inv}$.
The merging of two models represented with \textcircled{+} in the above equation,
and the seven model merging approaches detailed in Section \ref{3-2-2}
are applied to the merging process in our experiments.

If the norms of $\theta_{LLM}$ and $\theta_{bias}^{inv}$ are different,
we cannot examine the effect of the hyperparameter $\alpha$.
Therefore, we normalize the model weight $\theta_{bias}^{inv}$
to ensure that its norm is the same as that of $\theta_{LLM}$.

\subsubsection{Model Merging Algorithms} \label{3-2-2}
Our empirical experiments are conducted for seven merging techniques:
Linear \citep{Model_soups}, Karcher Mean \citep{grove1973conjugate},
SLERP \citep{slerp}, NuSLERP \citep{mergekit},
TIES \citep{TIES}, DELLA \citep{deep2024dellamerging}, and Nearswap \citep{mergekit}.
Utilizing these methods, we merge a bias-inverted model with a pre-trained LLM.

\textbf{Linear (Model Soups).}
\citet{Model_soups} proposed the most fundamental merging technique,
which adds and averages the weights of fine-tuned models with scaling parameters.
Through simple summation, it compactly integrates knowledge from multiple models,
yielding consistent performance at low cost.

\textbf{Karcher Mean.}
\citet{mergekit} introduced merging methods that compute the Karcher mean \citet{grove1973conjugate} on a Riemannian manifold to geometrically fuse models.
Unlike Linear merging, the Karcher Mean considers the curved geometry of the parameter manifold, preserving performance in non-Euclidean structures.

\textbf{SLERP.}
\citet{mergekit} presented the approach to interpolate the weight vectors of models along a great circle path on the hypersphere \citep{slerp}, preserving the curvature of parameter space.
SLERP constrains the path to the unit hypersphere, performing pairwise spherical interpolation.

\textbf{NuSLERP.}
\citet{mergekit} introduced an extension method of SLERP that assigns different interpolation ratios to each layer or tensor, enabling non-uniform spherical interpolation.
By weighting critical layers more heavily, it balances local expertise with global stability, achieving strong performance with simple rule-based settings.

\textbf{TIES.}
\citet{TIES} presented the method to merge models by extracting parameter differences that capture task-specific knowledge.
Sparsifying these differences, TIES is an algorithm to reduce interference and better preserve each model's strengths.

\textbf{DELLA.}
\citet{deep2024dellamerging} proposed the DELLA approach, which reduces interference by selectively pruning the less important task-specific parameter updates, using adaptive pruning with magnitude-aware rescaling.
It assigns higher keep probabilities to larger-magnitude parameters within each row, improving retention of important weights and matching original model performance.

\textbf{Nearswap.}
\citet{mergekit} proposed the merging method by strengthening the interpolation where the parameters are similar and weakening it when they differ.

\section{Experimental Setup}

\subsection{Models}
In order to compare different model architectures,
for our experiments, we selected three families of LLMs: GPT, LLAMA, and QWEN.

Specifically, the GPT family \citep{gpt2,gptneo} includes
GPT2-small, GPT2-medium, GPT2-large, GPT2-xl, and GPT-neo-2.7B.
The LLaMA family \citep{llama2,llama3} includes LLAMA-2-7B, LLAMA-3-8B, LLAMA-3.1-8B, LLAMA-3.2-1B, and LLAMA-3.2-3B.
Finally, the Qwen family \citep{qwen2} consists of QWEN2-0.5B, QWEN2-1.5B, and QWEN2-7B.

The models listed above are available from the Hugging Face repository,
and the URLs for all models are shown in Appendix~\ref{app:models}.

\subsection{Experimental Setup for Model Merging} \label{sec:4-2}
In merging models as described in Equation (\ref{eq:debias}), we vary the scaling factor \(\alpha\) from 0.1 to 0.5 in steps of 0.1.
The range of the $\alpha$ value is determined on the basis of the results of preliminary experiments (described in Appendix \ref{app:pre-experiments}).
Note that our model merging implementation is based on the mergekit toolkit \footnote{\url{https://github.com/arcee-ai/mergekit}.}, and the hyperparameters except for the scaling factor are set to the default values defined in the mergekit.

\paragraph{Continual Pre-Training Dataset.}
Following \citet{shirafuji-etal-2025-bias}, we use the StereoSet intrasentence dataset \citep{nadeem-etal-2021-stereoset} to construct bias-amplified models ($\theta_{bias}$).
Each sample in the original dataset contains a bias type (\textit{race}, \textit{profession}, \textit{gender}, or \textit{religion}), a sentence with one blank word, and three candidate words: stereotype, anti-stereotype, and meaningless.
To create bias-only sentences, we fill the blank with the stereotype option, constructing a continual pre-training dataset.

The computational resources for continual pre-training to create biased LLMs are described in Appendix \ref{app:compute},
and details of hyperparameter configurations are shown in Appendix \ref{app:hyperparameter}.

\subsection{Evaluation Dataset for Social Bias}
We evaluate social bias in LLMs using three benchmarks:
the Bias Benchmark for Question-Answering (BBQ) \citep{bbq},
the Bias in Open-Ended Language Generation Dataset (BOLD) \citep{bold},
and HONEST \citep{honest}.
The URLs of these datasets are listed in Appendix \ref{app:bias-data}.

\paragraph{BBQ.}
The BBQ dataset \citep{bbq} comprises approximately 58k templated question-answer pairs in nine social dimensions relevant to U.S. English speakers.
By contrasting ``underspecified'' with ``fully specified'' versions of each question,
it measures the extent to which models rely on stereotypical priors rather than explicit evidence.

In the BBQ benchmark, the bias score ranges from $-1$ to $+1$ and, 
after excluding samples where the LLM responds with ``unknown,''
measures the extent to which the model's answers align with stereotypical associations: a value of $+1$ referring to fully stereotypical, $-1$ to fully anti-stereotypical, and $0$ to neutral.

\paragraph{BOLD.}
The BOLD dataset \citep{bold}
contains 23,679 prompts, organized
into 43 demographic subgroups that cover occupation, gender, race, religion, and political ideology.

The generated text is classified by the \textit{regard} library \footnote{\url{https://huggingface.co/spaces/evaluate-measurement/regard}.}
into positive ($+1$), neutral ($0$), or negative ($-1$), and the absolute mean of the scores for each group is calculated as the bias score.
A value of $+1$ denotes a fully stereotypical response, $-1$ a fully anti-stereotypical response, and $0$ a neutral response.

\paragraph{HONEST.}
HONEST \citep{honest} is a multilingual, template- and lexicon-based benchmark to quantify harmful stereotypes in generated text. 
It comprises 420 identity–template prompts per language, and for each prompt, we collect the model's top-$K$ generated text and flag those containing \textsc{HurtLex} \citep{bassignana2018hurtlex} offensive terms
\footnote{\url{https://huggingface.co/spaces/evaluate-measurement/honest}.}.

Following \citet{honest}, we set $K=20$ and compute the bias score as the average proportion of completed assignments highlighted, where lower values indicate less bias.
We focus exclusively on English templates, since, as discussed in Section~\ref{sec:4-2}, the bias mitigated by model merging pertains only to the English bias held by the Americans.

\subsection{Evaluation Dataset: SuperGLUE}
To verify that the debiasing methods do not compromise performance on downstream tasks, 
we evaluate both the debiased and pre-trained LLMs on the SuperGLUE benchmark, 
which comprises eight tasks: BoolQ, CB, COPA, MultiRC, ReCoRD, RTE, WiC, and WSC.
All evaluations are conducted using the Language Model Evaluation Harness\footnote{\url{https://github.com/EleutherAI/lm-evaluation-harness}.}.

Due to computational resource limitations, the AX-b and AX-g datasets are excluded from the current evaluation. 
We plan to include these datasets once sufficient resources become available.

\section{Results and Discussion}
\subsection{Social Bias Evaluation}
The results of the bias scores on the BBQ, BOLD and HONEST datasets are shown in Figure~\ref{fig:bbq}, \ref{fig:bold}, and \ref{fig:honest}, respectively.
The detailed results are described in Appendix \ref{app:detail_results}.

\begin{figure*}[tp]
  \centering
  %--- サブ図 (a) ---
  \begin{subfigure}[b]{0.32\textwidth}
    \centering
    \includegraphics[width=\textwidth]{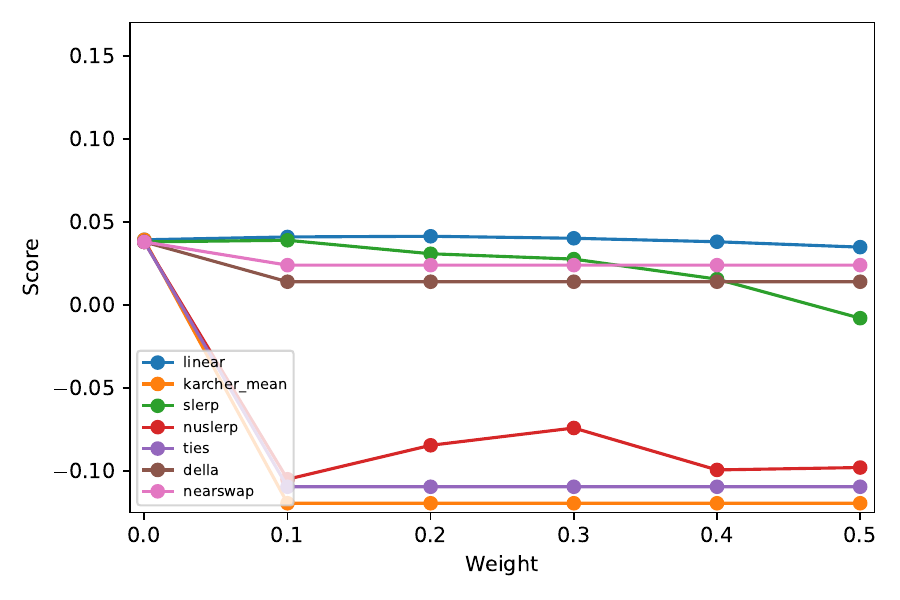}
    \subcaption{Avg. of GPT Family}
    \label{fig:bbq-gpt}
  \end{subfigure}
  \hfill
  %--- サブ図 (b) ---
  \begin{subfigure}[b]{0.32\textwidth}
    \centering
    \includegraphics[width=\textwidth]{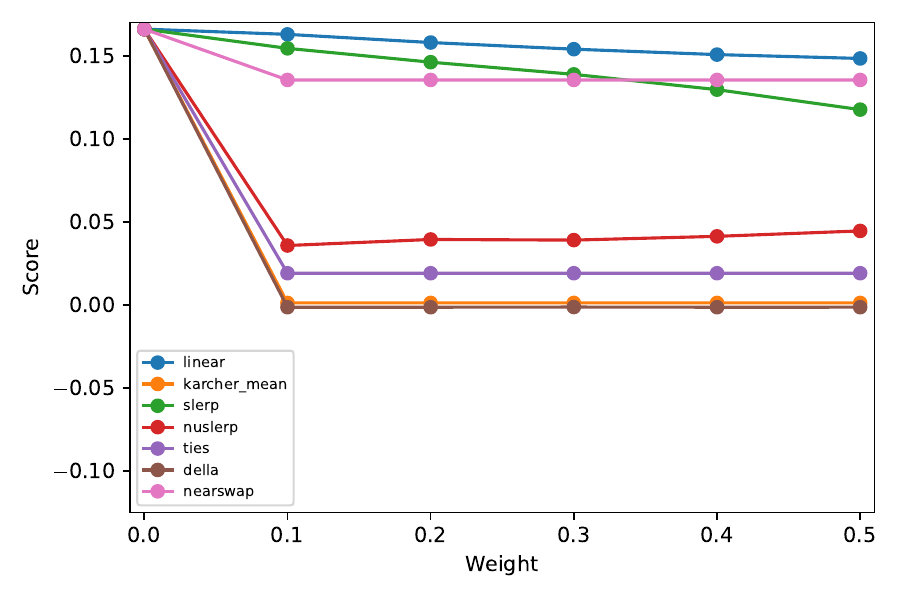}
    \subcaption{Avg. of LLAMA Family}
    \label{fig:bbq-llama}
  \end{subfigure}
  \hfill
  %--- サブ図 (c) ---
  \begin{subfigure}[b]{0.32\textwidth}
    \centering
    \includegraphics[width=\textwidth]{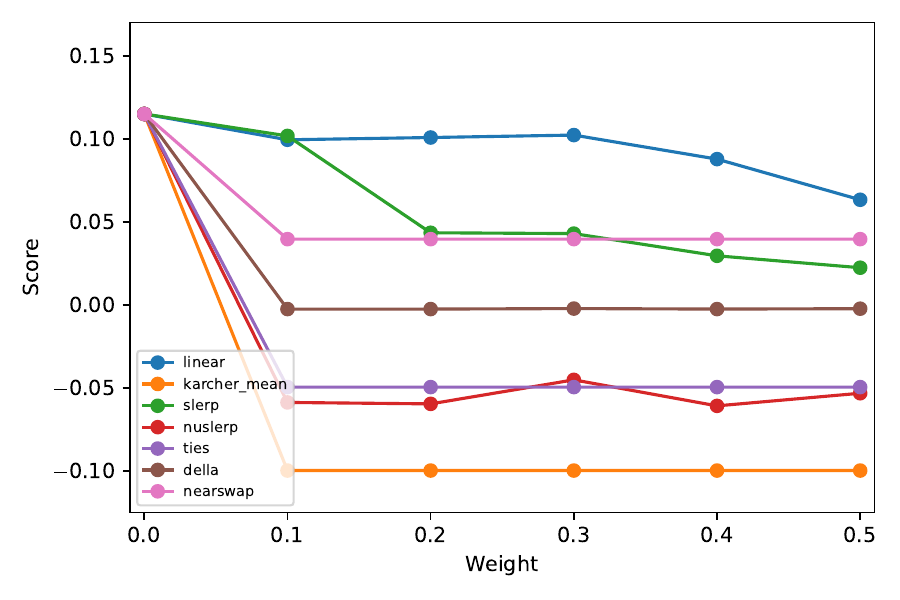}
    \subcaption{Avg. of QWEN Family}
    \label{fig:bbq-qwen}
  \end{subfigure}

  % 全体共通キャプション
  \caption{The BBQ evaluation results. Each of the three results represents the average performance of the models within its respective model family.
  The blue, orange, green, red, purple, brown, and pink lines correspond to the results for Linear, Karcher Mean, SLERP, NuSLERP, TIES, DELLA, and Nearswap, respectively.
  The scores of setting the weight $\alpha$ to zero are resulted using the pre-trained LLMs.}
  \label{fig:bbq}
\end{figure*}

\begin{figure*}[tp]
  \centering
  %--- サブ図 (a) ---
  \begin{subfigure}[b]{0.32\textwidth}
    \centering
    \includegraphics[width=\textwidth]{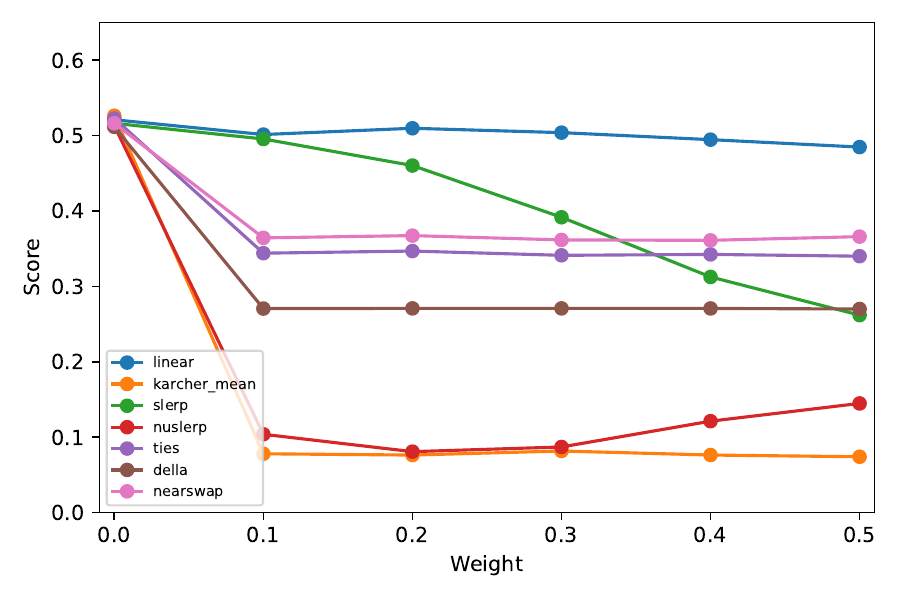}
    \subcaption{Avg. of GPT Family}
    \label{fig:bold-gpt}
  \end{subfigure}
  \hfill
  %--- サブ図 (b) ---
  \begin{subfigure}[b]{0.32\textwidth}
    \centering
    \includegraphics[width=\textwidth]{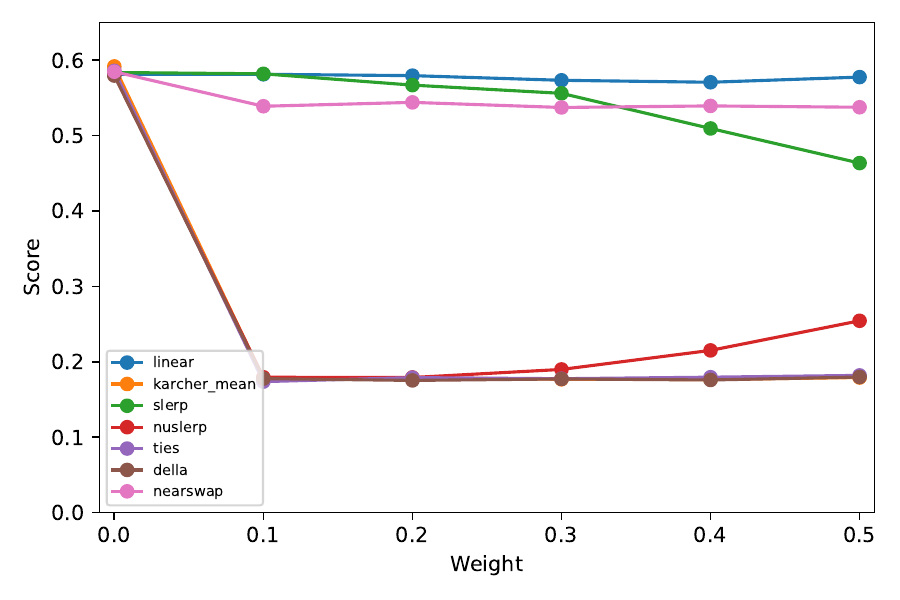}
    \subcaption{Avg. of LLAMA Family}
    \label{fig:bold-llama}
  \end{subfigure}
  \hfill
  %--- サブ図 (c) ---
  \begin{subfigure}[b]{0.32\textwidth}
    \centering
    \includegraphics[width=\textwidth]{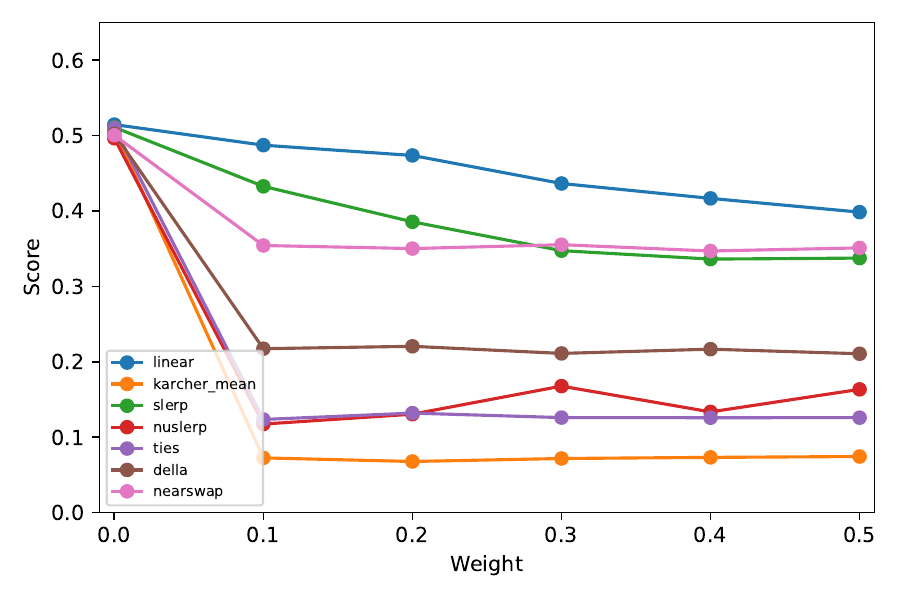}
    \subcaption{Avg. of QWEN Family}
    \label{fig:bold-qwen}
  \end{subfigure}

  % 全体共通キャプション
  \caption{The BOLD evaluation results. Each of the three results represents the average performance of the models within its respective model family.}
  \label{fig:bold}
\end{figure*}

\begin{figure*}[tp]
  \centering
  %--- サブ図 (a) ---
  \begin{subfigure}[b]{0.32\textwidth}
    \centering
    \includegraphics[width=\textwidth]{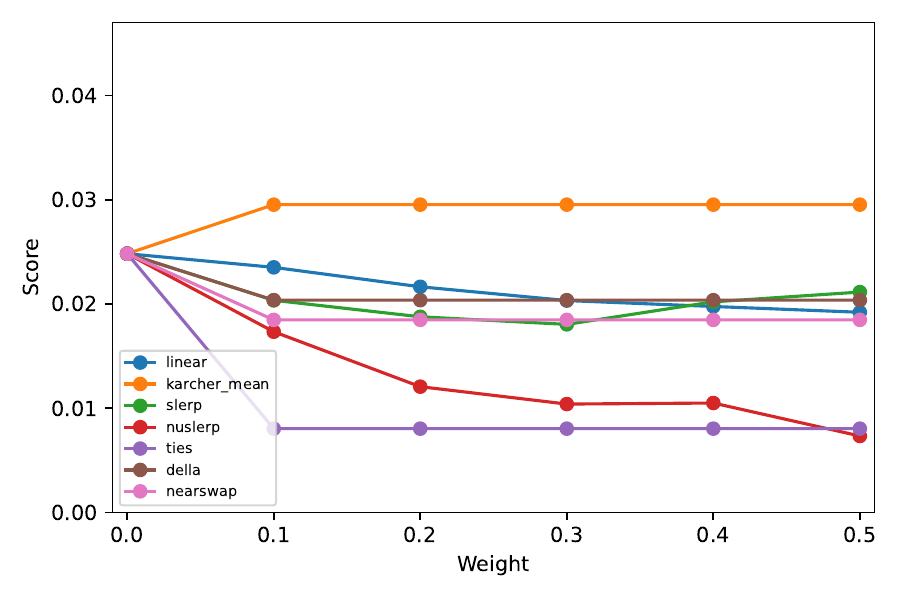}
    \subcaption{Avg. of GPT Family}
    \label{fig:honest-gpt}
  \end{subfigure}
  \hfill
  %--- サブ図 (b) ---
  \begin{subfigure}[b]{0.32\textwidth}
    \centering
    \includegraphics[width=\textwidth]{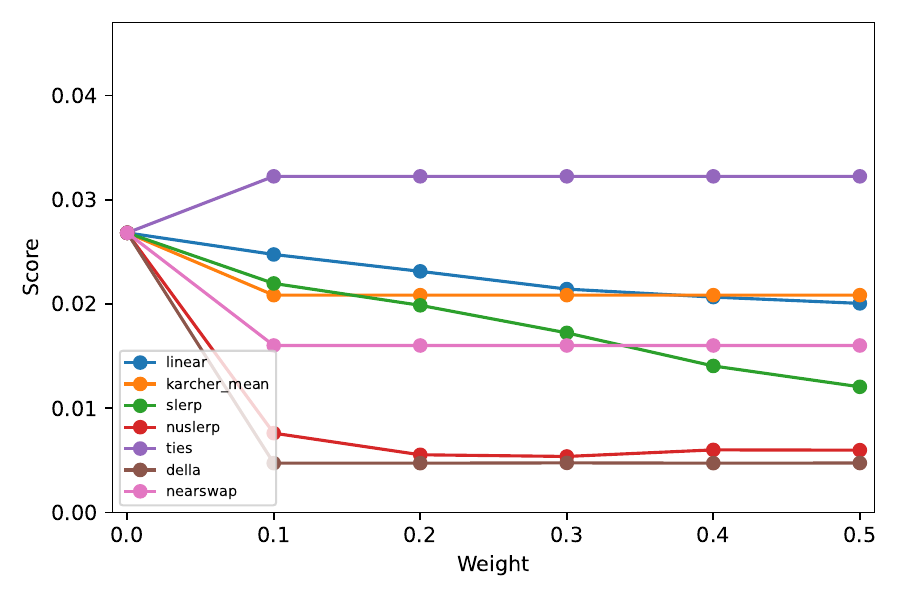}
    \subcaption{Avg. of LLAMA Family}
    \label{fig:honest-llama}
  \end{subfigure}
  \hfill
  %--- サブ図 (c) ---
  \begin{subfigure}[b]{0.32\textwidth}
    \centering
    \includegraphics[width=\textwidth]{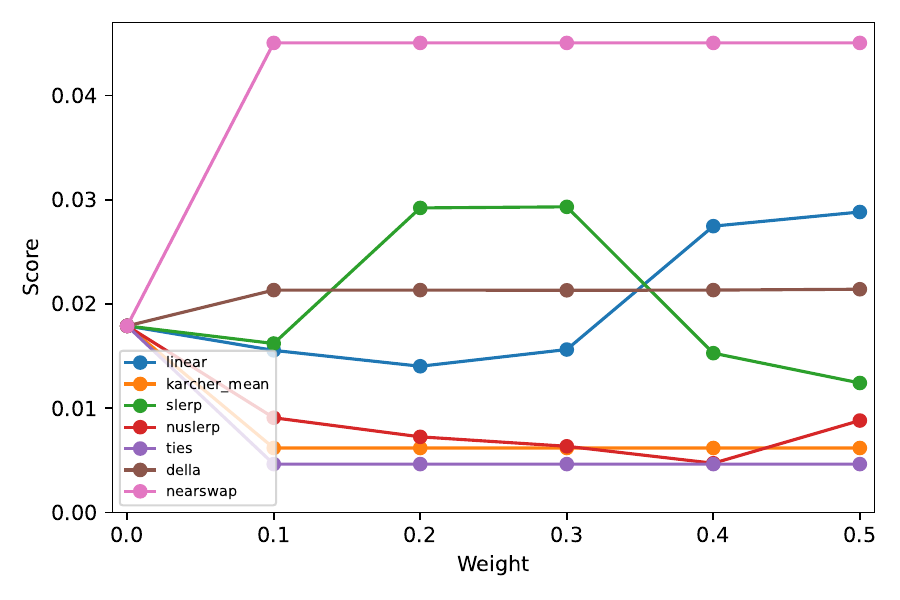}
    \subcaption{Avg. of QWEN Family}
    \label{fig:honest-qwen}
  \end{subfigure}

  % 全体共通キャプション
  \caption{The HONEST evaluation results. Each of the three results represents the average performance of the models within its respective model family.}
  \label{fig:honest}
\end{figure*}

\begin{figure*}[tp]
  \centering
  %--- サブ図 (a) ---
  \begin{subfigure}[b]{0.32\textwidth}
    \centering
    \includegraphics[width=\textwidth]{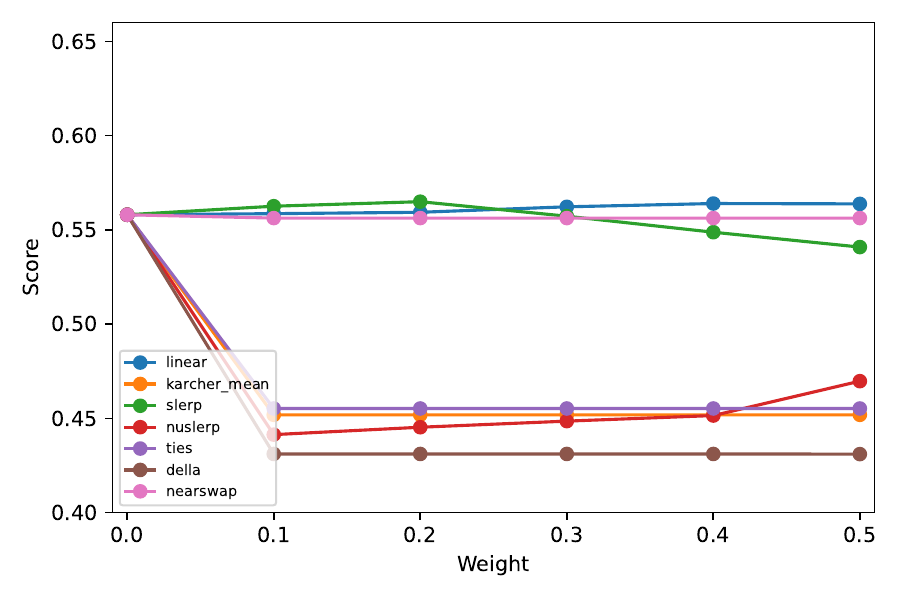}
    \subcaption{Avg. of GPT Family}
    \label{fig:sglue-gpt}
  \end{subfigure}
  \hfill
  %--- サブ図 (b) ---
  \begin{subfigure}[b]{0.32\textwidth}
    \centering
    \includegraphics[width=\textwidth]{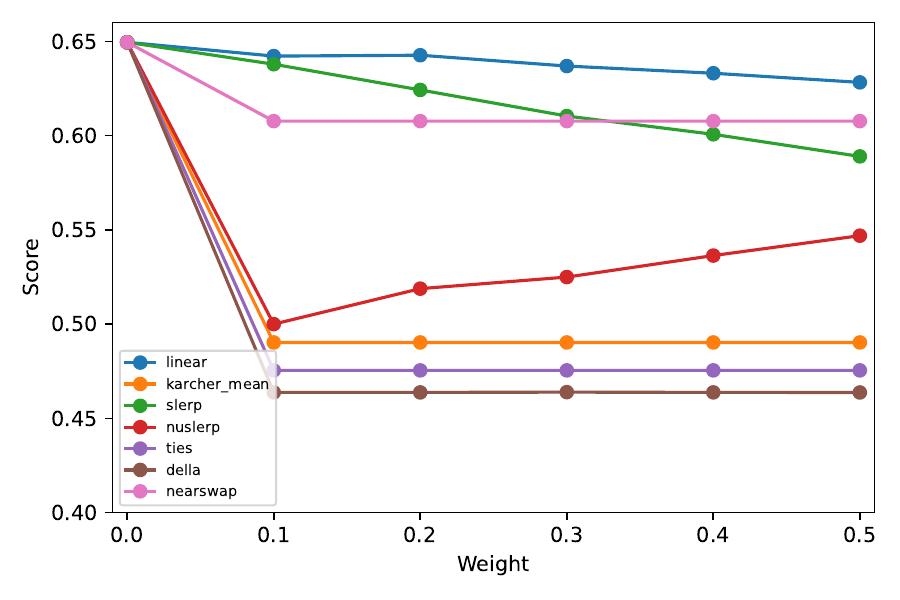}
    \subcaption{Avg. of LLAMA Family}
    \label{fig:sglue-llama}
  \end{subfigure}
  \hfill
  %--- サブ図 (c) ---
  \begin{subfigure}[b]{0.32\textwidth}
    \centering
    \includegraphics[width=\textwidth]{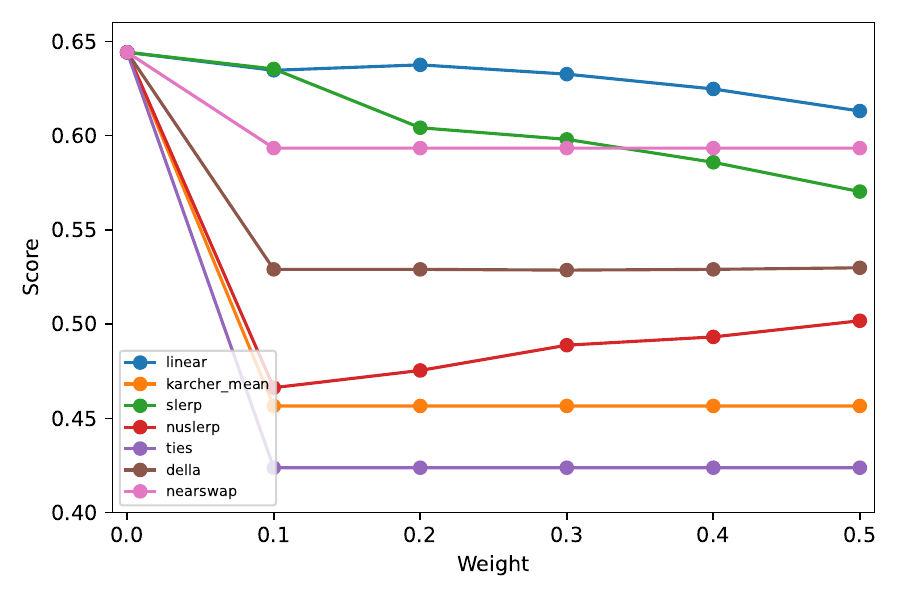}
    \subcaption{Avg. of QWEN Family}
    \label{fig:sglue-qwen}
  \end{subfigure}

  % 全体共通キャプション
  \caption{The SuperGLUE evaluation results. Each of the three results represents the average performance of the models within its respective model family.}
  \label{fig:sglue}
\end{figure*}

\paragraph{Overall Tendencies.}
 Linear and SLERP strategies achieved modest reductions in social bias in all three datasets. 
Nearswap further lowered the scores in most settings, with the notable exception of Qwen models in HONEST.

In contrast, Karcher Mean, NuSLERP, and TIES occasionally over-mitigated social biases, leading to anti-stereotypical outputs (e.g. -1.0 in GPT and Qwen in BBQ). 
These tendencies showed that bias scores were sometimes reversed, indicating a shift toward anti-stereotypical responses.

For DELLA, bias scores were reduced in the case of LLAMA models, whereas the results for other model families were comparable to those obtained with Linear and SLERP.

\paragraph{Impact of Model Architecture.}
Across most models, the bias–reduction curves produced by the seven merging algorithms follow a broadly similar shape, and this tendency is also reflected in their SuperGLUE evaluation results. In general, most methods produce an approximately linear decrease as the mixing factor varies ($\lambda \in [0,0.5]$).

However, some methods, such as NuSLERP, Karcher Mean, and occasionally Nearswap, exhibit irregular behavior in certain cases. Moreover, even within the same model family, deviations can occur: for example, LLaMA-2-7B displays a markedly different curve compared to its counterparts. This divergence is plausibly attributable to algorithmic differences between the LLaMA-2 and LLaMA-3 series.

Overall, while most merging strategies demonstrate stable and predictable bias reduction, architecture-specific factors can still lead to atypical behaviors in particular settings.

\paragraph{Model Parameters.}
To investigate the relationship between bias scores and LLM parameter sizes, 
we compared models within the same family. 
Bias scores in BBQ for individual models are provided in the Appendix~\ref{app:detail_results}.

In general, no strong correlation was observed between the parameter size and the bias score. 
Although some models (e.g., GPT-2-medium, LLaMA-2-7B, Qwen2-0.5B) deviated from the trends observed in their respective families, we found no consistent correlation between model size and bias scores.

\begin{table*}[t]
    \centering
    \begin{tabular}{l|r|rrrrrrrr}
         Methods & \multicolumn{1}{c|}{$\alpha$} & \multicolumn{1}{c}{BoolQ} & \multicolumn{1}{c}{CB} & \multicolumn{1}{c}{COPA} & \multicolumn{1}{c}{MultiRC} & \multicolumn{1}{c}{ReCoRD} & \multicolumn{1}{c}{RTE} & \multicolumn{1}{c}{WiC} & \multicolumn{1}{c}{WSC} \\
         \hline\hline
         pre-trained & \multicolumn{1}{c|}{--} & 0.683 & 0.501 & 0.786 & 0.487 & 0.854 & 0.598 & 0.504 & 0.490 \\\hline
         linear & 0.1 & 0.688 & 0.495 & 0.777 & 0.491 & 0.854 & 0.595 & 0.502 & 0.464 \\
                & 0.5 & 0.683 & 0.477 & 0.748 & 0.504 & 0.823 & 0.589 & 0.503 & 0.472 \\\hline
         karcher-mean & 0.1 & \cellcolor{LightRed}0.511 & 0.382 & \cellcolor{LightRed}0.592 & 0.509 &
                \cellcolor{LightRed}0.252 & 0.517 & 0.495 & 0.484\\
                 & 0.5 & \cellcolor{LightRed}0.511 & 0.382 & \cellcolor{LightRed}0.592 & 0.509 & 
                 \cellcolor{LightRed}0.252 & 0.517 & 0.495 & 0.484 \\\hline
         slerp & 0.1 & 0.686 & 0.497 & 0.766 & 0.500 & 0.848 & 0.594 & 0.505 & 0.470 \\
                 & 0.5 & 0.656 & 0.426 & 0.660 & 0.520 & 0.755 & 0.563 & 0.501 & 0.448 \\\hline
         nuslerp & 0.1 & \cellcolor{LightRed}0.505 & 0.357 & 0.584 & 0.491 & 
                \cellcolor{LightRed}0.304 & 0.517 & 0.499 & 0.500 \\
                  & 0.5 & \cellcolor{LightRed}0.507 & \cellcolor{LightRed}0.352 & \cellcolor{LightRed}0.625 & 0.489 &
                  \cellcolor{LightRed}0.532 & 0.528 & 0.498 & 0.524 \\\hline
         ties    & 0.1 & \cellcolor{LightRed}0.511 & \cellcolor{LightRed}0.331 & \cellcolor{LightRed}0.570 & 0.500 &
                \cellcolor{LightRed}0.214 & 0.521 & 0.500 & 0.499 \\
                 & 0.5 & \cellcolor{LightRed}0.511 & \cellcolor{LightRed}0.331 & \cellcolor{LightRed}0.570 & 0.500 &
                 \cellcolor{LightRed}0.214 & 0.521 & 0.500 & 0.499 \\\hline
         della & 0.1 & 0.565 & \cellcolor{LightRed}0.277 & \cellcolor{LightRed}0.573 & 0.513 &
                \cellcolor{LightRed}0.272 & 0.538 & 0.501 & 0.490 \\
                 & 0.5 & 0.565 & \cellcolor{LightRed}0.279 & \cellcolor{LightRed}0.573 & 0.512 &
                 \cellcolor{LightRed}0.272 & 0.537 & 0.501 & 0.491 \\\hline
         nearswap & 0.1 & 0.666 & 0.446 & 0.713 & 0.508 & 0.788 & 0.588 & 0.510 & 0.459 \\
                  & 0.5 & 0.666 & 0.446 & 0.713 & 0.508 & 0.788 & 0.588 & 0.510 & 0.459 \\\hline
         %breadcrumbs & \multicolumn{1}{c|}{--} & 0.553 & \cellcolor{LightRed}0.243 & \cellcolor{LightRed}0.568 & 0.527 &
         %\cellcolor{LightRed}0.271 & 0.536 & 0.501 & 0.473 \\\hline
    \end{tabular}
    \caption{SuperGLUE evaluation scores on each task with pre-trained LLMs and the debiased LLMs by the model merging methods,
    setting a scaling factor $\alpha$ to 0.1 or 0.5.
    Results highlighted in red indicate scores that are more than 15\% lower than those of the pre-trained LLM.}
    \label{tab:sglue_detail}
\end{table*}

\subsection{SuperGLUE Evaluation} \label{sec5-2}
The aggregated SuperGLUE results are shown in Figure~\ref{fig:sglue}.

Two main observations emerge from the results:
(i) increasing the scaling factor consistently decreases SuperGLUE scores in most cases; and
(ii) Linear, SLERP, and Nearswap preserve downstream performance, and the remaining four techniques reduce average scores by more than 10\%.

To identify which abilities were most affected, Table~\ref{tab:sglue_detail} reports task-wise scores averaged over all LLMs. 
Relative to the three stable methods, the other approaches substantially impair performance on  
ReCoRD ($\downarrow$~50--60\%),  
BoolQ ($\downarrow$~15--20\%), 
COPA ($\downarrow$~15--20\%),  
and CB ($\downarrow$~10--20\%),  
while leaving the other SuperGLUE tasks largely unaffected.

Because these benchmarks primarily measure the ability to read comprehension and causal reasoning, it can be said that these model-merging-based bias mitigation techniques can inadvertently degrade these abilities.
Even the more stable methods (Linear, NuSLERP, and Nearswap) show minor decreases of $\downarrow$~ 2--3\%, $\downarrow$~ 2--10\%, and $\downarrow$~7\%, respectively.
Furthermore, in all methods, the larger $\alpha$ becomes, i.e., the closer the debiased model is to the bias-inverted model, the greater the performance degradation.

Our findings are consistent with the results of the task vector-based approach of \citet{shirafuji-etal-2025-bias}, which also reported that the debiased models maintain the general precision of the GLUE, but suffer substantial losses in CoLA (over $\downarrow$~ 50\%), a task that evaluates grammatical acceptability.

In contrast, some existing debiasing studies \citep{lutz-etal-2024-local} based on model merging have demonstrated the performance of the downstream tasks of debiased LLMs using scores from NLI benchmarks. 
Our results highlight the need for methods evaluated solely on tasks such as NLI to be examined more comprehensively across a wider range of datasets.

\subsection{Which Merging Algorithm is the Most Accurate for Social Bias Mitigation?}

SuperGLUE results indicate that, except for Linear, SLERP, and Nearswap, the other merging techniques substantially degrade the causal reasoning capabilities of LLMs (Section~\ref{sec5-2}). Consequently, these methods are unsuitable for reliable bias mitigation.

Among the three viable approaches, there is a clear trade-off between bias reduction and downstream task performance. SLERP and Nearswap achieve the largest reductions in bias but incur an average SuperGLUE decline of approximately 5\%. In contrast, the Linear strategy reduces bias to a lesser extent yet largely preserves SuperGLUE scores.

In particular, SLERP with moderate interpolation weights ($\alpha = 0.2$–$0.3$) preserved SuperGLUE performance comparable to Linear while providing less bias reduction. Therefore, we recommend SLERP at $\alpha=0.2-0.3$ as the most effective compromise.

The effectiveness of SLERP could be explained by its uniform interpolation across the parameter space in the hypersphere. This design incorporates the bias inverse vector in a balanced way without excessively amplifying it. In contrast, the other interpolating approach (NuSLERP) performed normalization at the layer or tensor level, substantially affecting its SuperGLUE scores.

This difference in accuracy arises from the fact that SLERP merges parameters across all layers as a whole, while NuSLERP performs the merging at the level of individual layers. In other words, SLERP preserves the global balance of interpolation and maintains a consistent meaning of $\alpha$ throughout the model, while NuSLERP rescales each layer separately, which amplifies local variations and leads to unstable behavior when all layers are merged simultaneously.

These findings suggest that, unlike SLERP, most recent model-merging methods cannot be directly applied for bias mitigation without risking substantial losses in reasoning performance.

\section{Conclusions and Future Works}
This work presented the first comprehensive study of how seven model-merging algorithms influence social bias in LLM.
By evaluating 13 models spanning the GPT, LLaMA, and Qwen families on three social bias datasets and the SuperGLUE benchmark, we revealed a trade-off between fairness and utility.

Linear, SLERP, and Nearswap consistently mitigated stereotypical tendencies across all architectures, whereas Karcher Mean, NuSLERP, TIES, and DELLA often reduced social bias excessively, resulting in LLMs that exhibit anti-stereotypical behavior.
Among the seven methods, SLERP with moderate interpolation weights ($\alpha = 0.2\text{--}0.3$) proved to be the most balanced approach, achieving a greater bias reduction than Linear while maintaining downstream accuracy.

Our analysis also revealed that bias reduction patterns were broadly consistent across architectures, with the notable exception of LLaMA2-7B.
Trends with respect to the scaling factor $\alpha$ also remained stable regardless of model size, suggesting that parameter scale alone does not alter the fundamental dynamics of merging.

In addition, the four methods (Karcher Mean, NuSLERP, TIES, and DELLA) substantially degraded performance on tasks requiring reading comprehension and commonsense or causal reasoning, such as ReCoRD, COPA, CB, and BoolQ in the SuperGLUE benchmark.
Some existing debiasing methods based on model merging have demonstrated their debiased LLMs' downstream-task performance using scores from NLI benchmarks.
However, we revealed that it is also essential to verify accuracy on tasks for reading comprehension and commonsense / causal reasoning.

In future work, to preserve these capabilities of debiased LLMs, we plan to jointly merge models specialized for these tasks during bias mitigation via model merging.

%\section*{Limitations}
%XXX

\section*{Ethics Statement}
\citet{Navigli} define \textit{bias} in natural language processing as
``prejudices, stereotypes, and discriminatory attitudes against certain groups of people.'' 
We adopt this definition throughout this paper.

For simplicity, we use the term ``bias'' to refer to both stereotypes and biases, while acknowledging that they are distinct concepts. 
We also recognize that the stereotypical data (StereoSet) used in our experiments reflect the biases of U.S. residents \citep{nadeem-etal-2021-stereoset}.

Our work specifically addressed bias mitigation in LLMs by leveraging stereotypes. 
Biases arise when concepts that should not be associated with particular social groups are unfairly linked. 
If LLM systems exhibit such biases, they may leave a negative impression on users. 
Our study examines the applicability of a task-arithmetic approach to mitigate bias, with the aim of reducing LM bias using the proposed methods.

We recognize the importance of maintaining an objective position. 
Therefore, we emphasize that the content of this study is not influenced by any political positions, stereotypes, or biases of the authors. 
Our research is guided by the ethical principle of fairness in scientific inquiry and seeks to make constructive and responsible contributions to the development of AI technologies.

\bibliography{custom}

\newpage

\appendix

\section{Model List} \label{app:models}
In this section, we show the model list and these URLs available in HuggingFace repositories.

\begin{itemize}
    \item GPT2-small: \url{https://huggingface.co/openai-community/gpt2},
    \item GPT2-medium: \url{https://huggingface.co/openai-community/gpt2-medium},
    \item GPT2-large: \url{https://huggingface.co/openai-community/gpt2-large},
    \item GPT2-xl: \url{https://huggingface.co/openai-community/gpt2-xl},
    \item GPT-neo-2.7B: \url{https://huggingface.co/EleutherAI/gpt-neo-2.7B},

    \item LLAMA-2-7B: \url{https://huggingface.co/meta-llama/Llama-2-7b-hf},
    \item LLAMA-3-8B: \url{https://huggingface.co/meta-llama/Meta-Llama-3-8B-hf},
    \item LLAMA-3.1-8B: \url{https://huggingface.co/meta-llama/Llama-3.1-8B},
    \item LLAMA-3.2-1B: \url{https://huggingface.co/meta-llama/Llama-3.2-1B},
    \item LLAMA-3.2-3B: \url{https://huggingface.co/meta-llama/Llama-3.2-3B},

    \item QWEN2-0.5B: \url{https://huggingface.co/Qwen/Qwen2-0.5B},
    \item QWEN2-1.5B: \url{https://huggingface.co/Qwen/Qwen2.5-1.5B},
    \item QWEN2-7B: \url{https://huggingface.co/Qwen/Qwen2-7B}.
\end{itemize}

\section{Preliminary Experiments} \label{app:pre-experiments}

\begin{figure}[t]
  \centering
    \centering
    \includegraphics[width=\columnwidth]{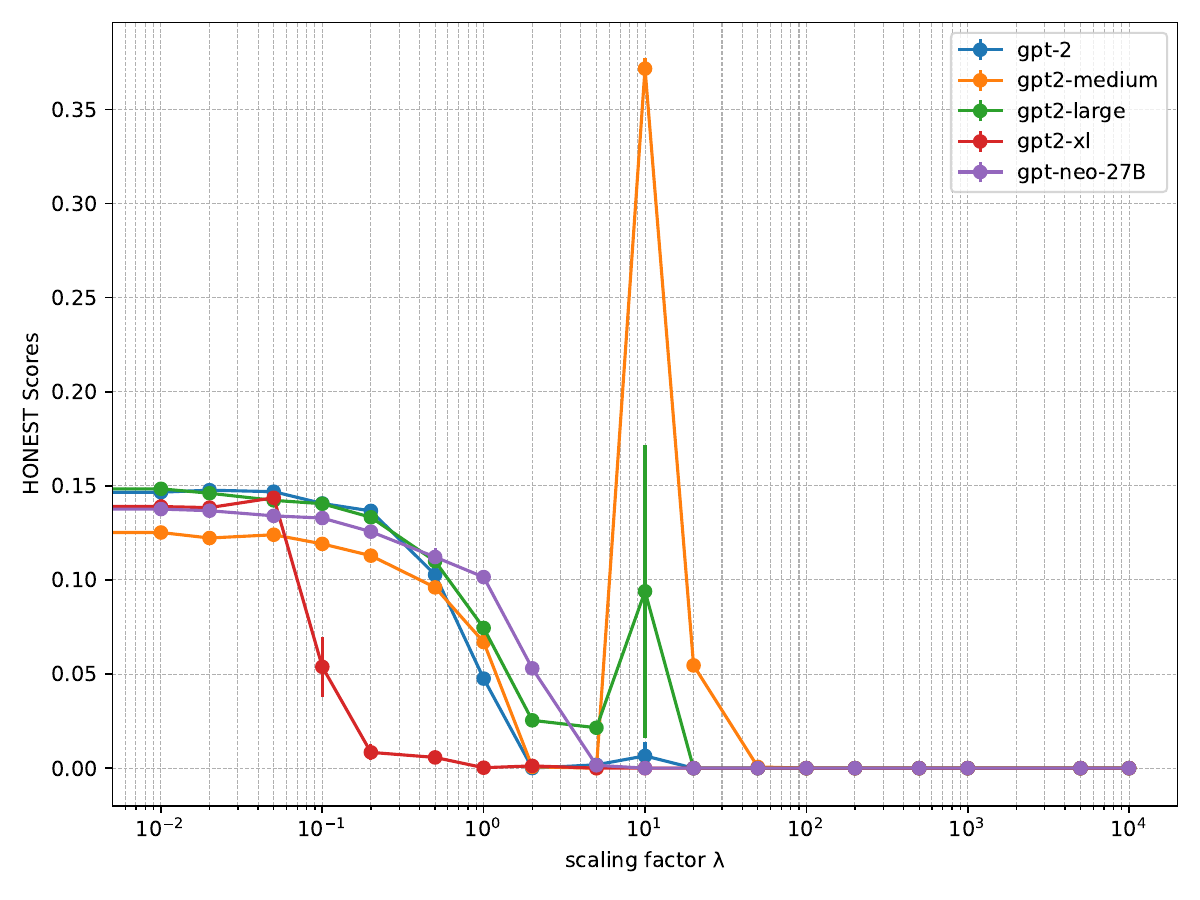}
    \caption{HONEST evaluation results of GPT model families on Preliminary experiments.}
    \label{fig:honest_gpt_app}
\end{figure}

\begin{table}[t]
  \centering
  \scalebox{1}{
  \begin{tabular}{l|rr}
    \hline
    Methods
    & \multicolumn{1}{c}{cola} & \multicolumn{1}{c}{avg.} \\
    \hline\hline
    GPT2-small
    & \textbf{0.449} & \textbf{0.760} \\
    ~w/ Bias Vector ($\alpha=0.1$)
    & 0.396 & 0.754 \\
    ~w/ Bias Vector ($\alpha=0.2$)%cola, rte
    & 0.440 & 0.759 \\
    ~w/ Bias Vector ($\alpha=0.5$)
    & 0.362 & 0.754 \\
    ~w/ Bias Vector ($\alpha=1$)
    & 0.050 & 0.702 \\
    ~w/ Bias Vector ($\alpha=2$)
    & 0.012 & 0.705 \\
    ~w/ Bias Vector ($\alpha=5$)
    & 0.000 & 0.669 \\
    ~w/ Bias Vector ($\alpha=10$)
    & 0.016 & 0.590 \\
    \hline
    \end{tabular}
  }
  \caption{GLUE evaluation results of GPT2-small on the preliminary experiments.}
  \label{tab-glue_detail_gpt2}
\end{table}

This section describes preliminary experiments conducted to narrow down the appropriate range for the hyperparameter $\alpha$.

We first evaluated GPT-based models using HONEST and GLUE \citep{wang2018glue} in advance to determine the effective range of $\alpha$.
In this experiment, $\alpha$ was set to 0.1, 0.2, 0.5, 1, 2, 5, and 10, and the model merging method followed the approach proposed by \citet{shirafuji-etal-2025-bias}.

The experimental results of HONEST with $K=20$ are shown in Figure \ref{fig:honest_gpt_app}, and the results of the evaluation of GPT2-small in GLUE are presented in Table \ref{tab-glue_detail_gpt2}.
From the evaluation, we found that for values of $\alpha$ around 5, bias was nearly eliminated for all models.
However, in certain downstream tasks (COLA), performance began to gradually degrade from $\alpha=0.5$ and dropped to almost zero at $\alpha=1$.

Based on these results, the main experiments in this paper restrict $\alpha$ to the range of 0.1 to 0.5.

\section{Computational Environment}\label{app:compute}
All LLM training for the stereotypical bias experiments was performed on AWS \texttt{p4d.24xlarge} instances, each equipped with eight NVIDIA H100 GPUs. 
Models with up to 3~billion parameters were trained on four H100 GPUs, while larger models used all eight.

For the evaluation experiments on SuperGLUE, BBQ, BOLD, and HONEST, all runs -- except those for the GPT-based model family -- were conducted on NVIDIA H100 GPUs:
models with up to 3~billion parameters used a single GPU for inference and scoring, and larger models were allocated two GPUs. 
 GPT-based models were evaluated on an NVIDIA Quadro RTX~8000.

\begin{table}[t]
  \centering
  \scalebox{1}{
  \begin{tabular}{l|cc}
    \hline
    Model & lr & scheduler \\
    \hline\hline
    GPT2-small & 3e-5 & linear \\
    GPT2-medium & 3e-5 & linear \\
    GPT2-large & 2e-5 & linear \\
    GPT2-xl & 1e-5 & linear \\
    GPT2-neo-2.7B & 1e-5 & linear \\
    LLAMA-2-7B & 1e-5 & cosine \\
    LLAMA-3-8B & 1e-5 & cosine \\
    LLAMA-3.1-8B & 1e-5 & cosine \\
    LLAMA-3.2-1B & 2e-5 & cosine \\
    LLAMA-3.2-3B & 1e-5 & cosine \\
    QWEN2-0.5B & 1e-4 & cosine \\
    QWEN2-1.5B & 2e-5 & cosine \\
    QWEN2-7B & 1e-5 & cosine \\

    \hline
    \end{tabular}
  }
  \caption{Hyperparameter configurations for LLM training. ``lr'' denotes the learning rate, and ``scheduler'' indicates the learning rate scheduling strategy.}
  \label{tab-ex_setup}
\end{table}

\section{Hyperparameter Configurations} \label{app:hyperparameter}

The experimental setup for continual learning is designed as follows.
We utilize the HuggingFace AutoModelForCausalLM library for model training. To reduce GPU memory consumption, the maximum sequence length (max\_length) is set to 512, the batch size is set to 64.
Training is carried out for 30 epochs with a weight decay of 0.01 and a warm-up ratio of 0.1.

The hyperparameters specific to each model, namely the learning rate and the learning rate scheduler, are described in Table~\ref{tab-ex_setup}.

Note that the scheduler was set to linear for the GPT family but cosine for the other models, since we followed the configuration of \citet{shirafuji-etal-2025-bias}, which established the linear scheduler as the default choice for GPT.

\section{List of Evaluation Datasets} \label{app:bias-data}
The URLs of the social bias evaluation datasets are listed as follows:
\begin{itemize}
    \item BBQ: \url{https://huggingface.co/datasets/heegyu/bbq};
    \item BOLD: \url{https://huggingface.co/datasets/AmazonScience/bold};
    \item HONEST: \url{https://huggingface.co/datasets/MilaNLProc/honest}.
\end{itemize}

\section{Each LLM Result on SuperGLUE, BBQ, BOLD, and HONEST} \label{app:detail_results}
\begin{figure*}[htbp]
  \centering
  % 1行目
  \begin{subfigure}[b]{0.45\textwidth}
    \centering
    \includegraphics[width=\linewidth]{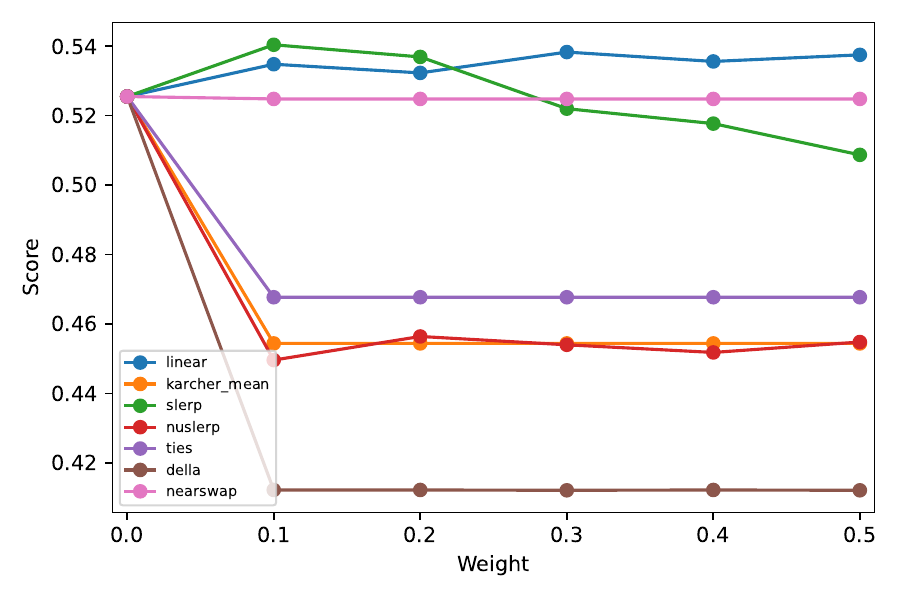}
    \caption{GPT2-small}
  \end{subfigure}
  \hfill
  \begin{subfigure}[b]{0.45\textwidth}
    \centering
    \includegraphics[width=\linewidth]{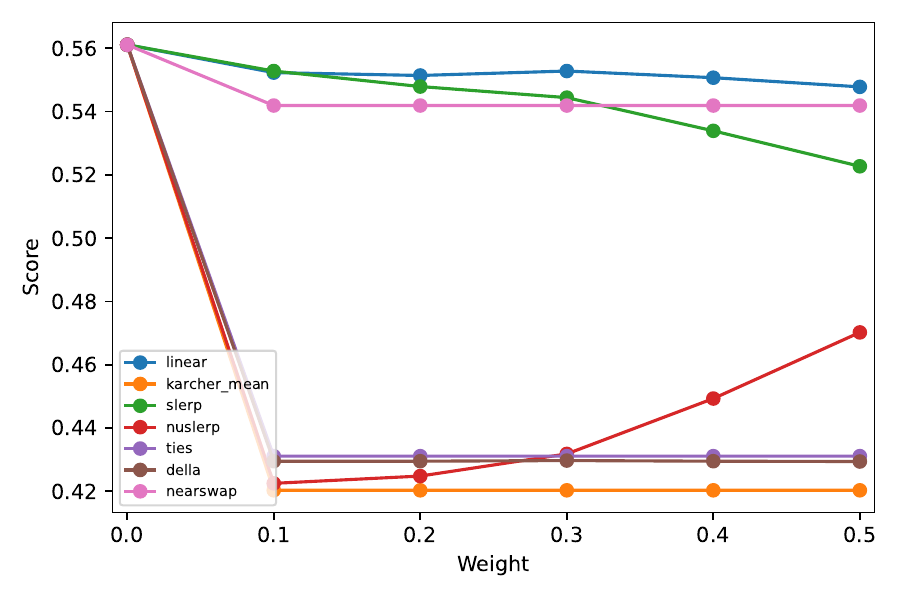}
    \caption{GPT2-medium}
  \end{subfigure}

  \vspace{1em}
  % 2行目
  \begin{subfigure}[b]{0.45\textwidth}
    \centering
    \includegraphics[width=\linewidth]{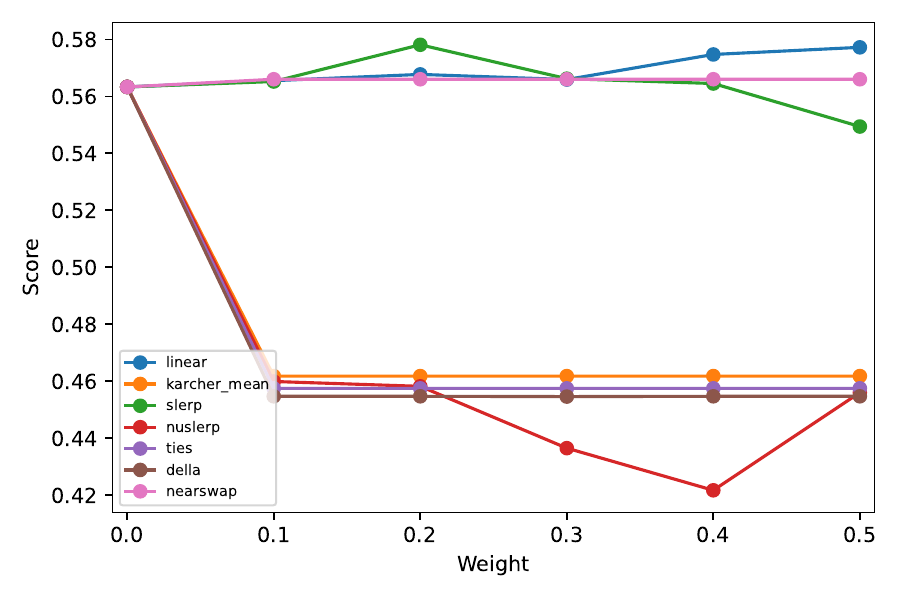}
    \caption{GPT2-large}
  \end{subfigure}
  \hfill
  \begin{subfigure}[b]{0.45\textwidth}
    \centering
    \includegraphics[width=\linewidth]{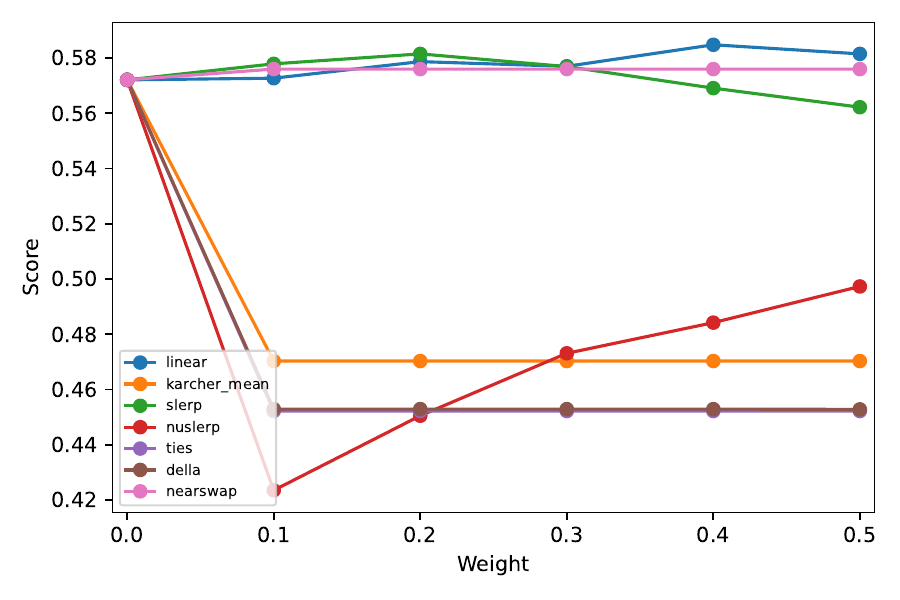}
    \caption{GPT2-XL}
  \end{subfigure}
  
  \vspace{1em}
  % 3行目
  \begin{subfigure}[b]{0.45\textwidth}
    \centering
    \includegraphics[width=\linewidth]{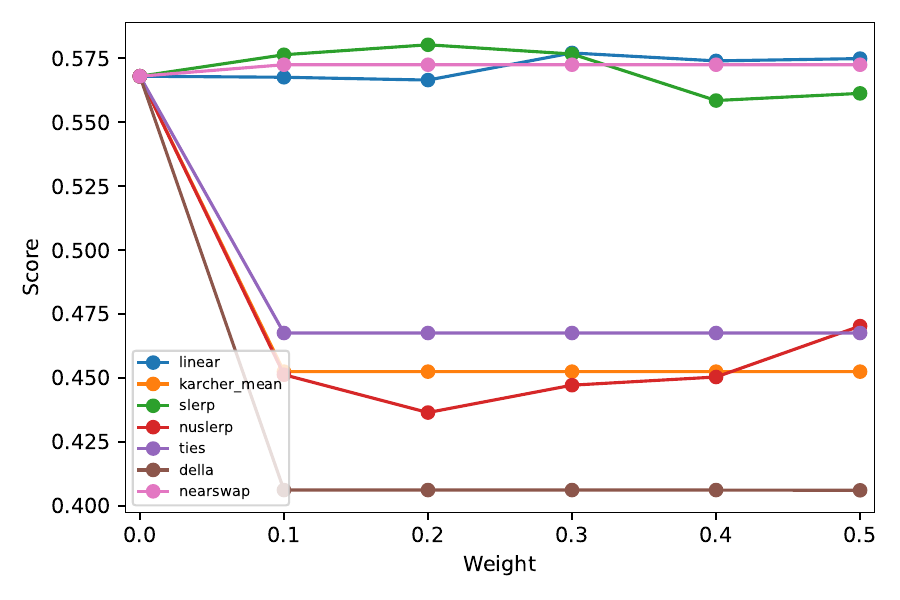}
    \caption{GPT-Neo-2.7B}
  \end{subfigure}

  \caption{The SuperGLUE evaluation results in GPT models.
  The blue, orange, green, red, purple, brown, and pink lines correspond to the results for Linear, Karcher Mean, SLERP, NuSLERP, TIES, DELLA, and Nearswap, respectively.
  The scores of setting the weight $\alpha$ to zero are resulted using the pre-trained LLMs.}
  \label{fig:app-sglue-gpt}
\end{figure*}

\begin{figure*}[htbp]
  \centering
  % 1行目
  \begin{subfigure}[b]{0.45\textwidth}
    \centering
    \includegraphics[width=\linewidth]{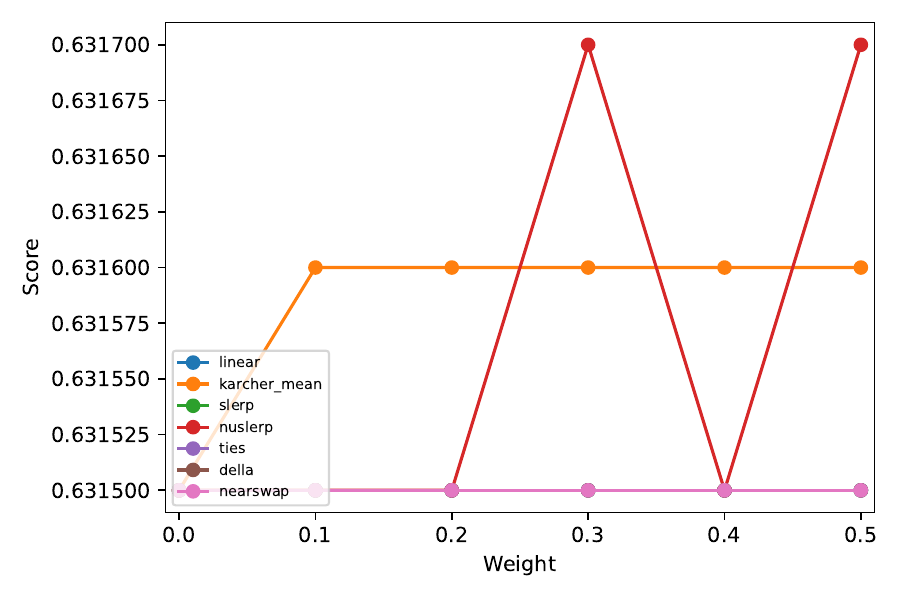}
    \caption{LLAMA2-7B}
  \end{subfigure}
  \hfill
  \begin{subfigure}[b]{0.45\textwidth}
    \centering
    \includegraphics[width=\linewidth]{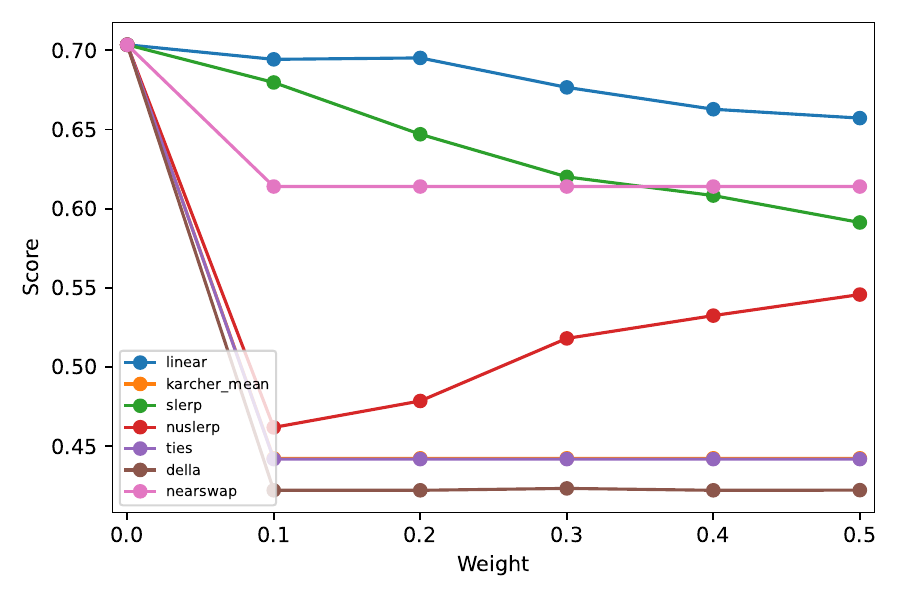}
    \caption{LLAMA-3.1-8B}
  \end{subfigure}

  \vspace{1em}
  % 2行目
  \begin{subfigure}[b]{0.45\textwidth}
    \centering
    \includegraphics[width=\linewidth]{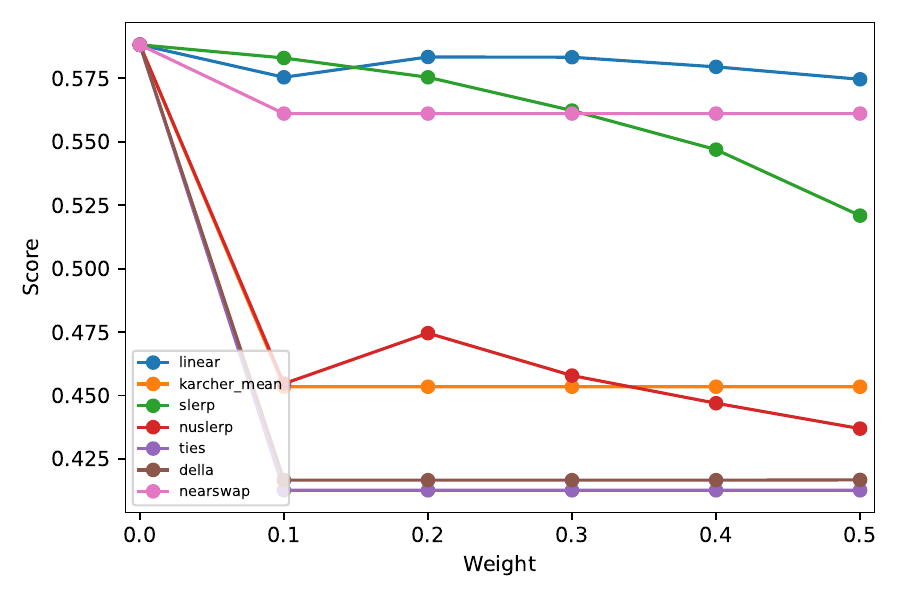}
    \caption{LLAMA-3.2-1B}
  \end{subfigure}
  \hfill
  \begin{subfigure}[b]{0.45\textwidth}
    \centering
    \includegraphics[width=\linewidth]{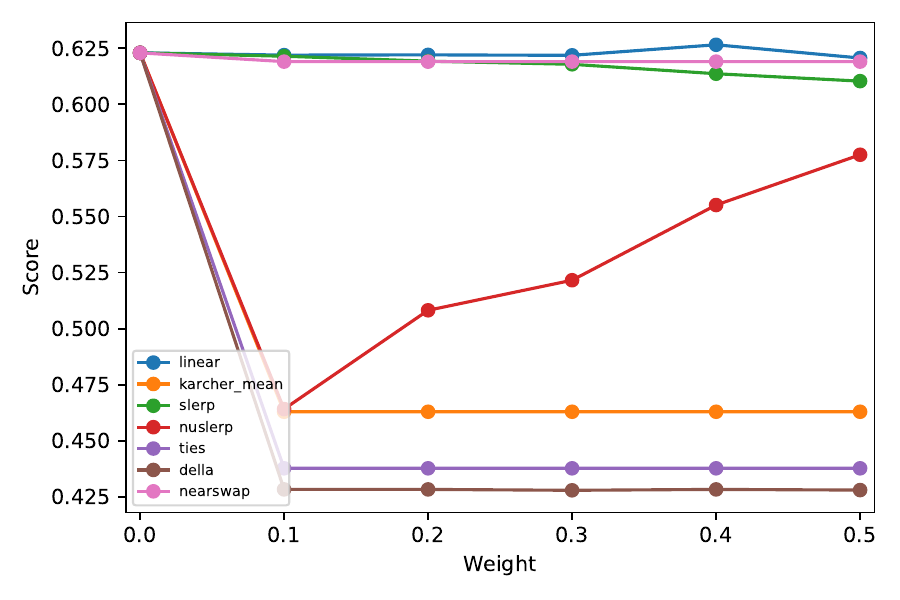}
    \caption{LLAMA-3.2-3B}
  \end{subfigure}
  \caption{The SuperGLUE evaluation results in LLAMA models.}
  \label{fig:app-sglue-llama}
\end{figure*}

\begin{figure*}[htbp]
  \centering
  \vspace{1em}
  % 3行目
  \begin{subfigure}[b]{0.4\textwidth}
    \centering
    \includegraphics[width=\linewidth]{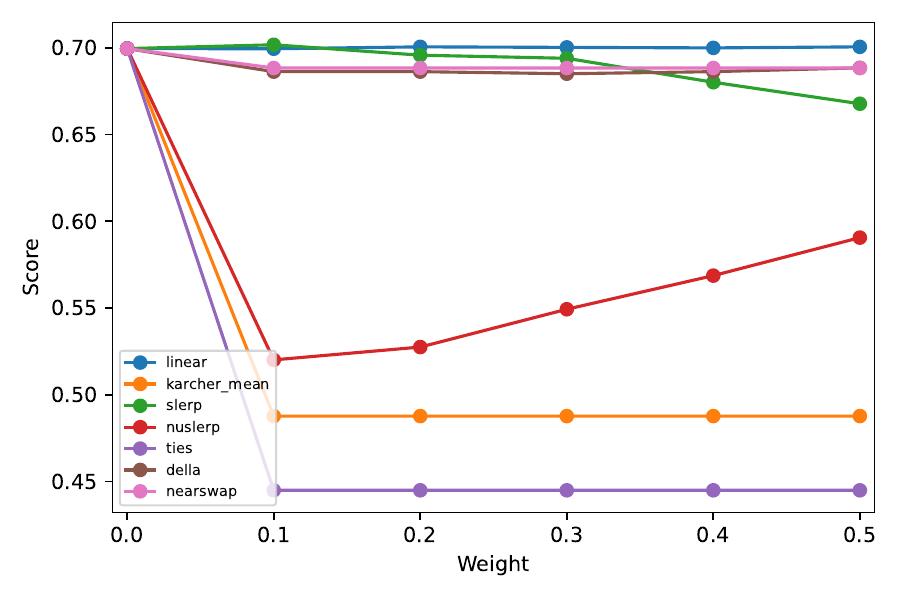}
    \caption{QWEN2-0.5B}
  \end{subfigure}
  \hfill
  \begin{subfigure}[b]{0.4\textwidth}
    \centering
    \includegraphics[width=\linewidth]{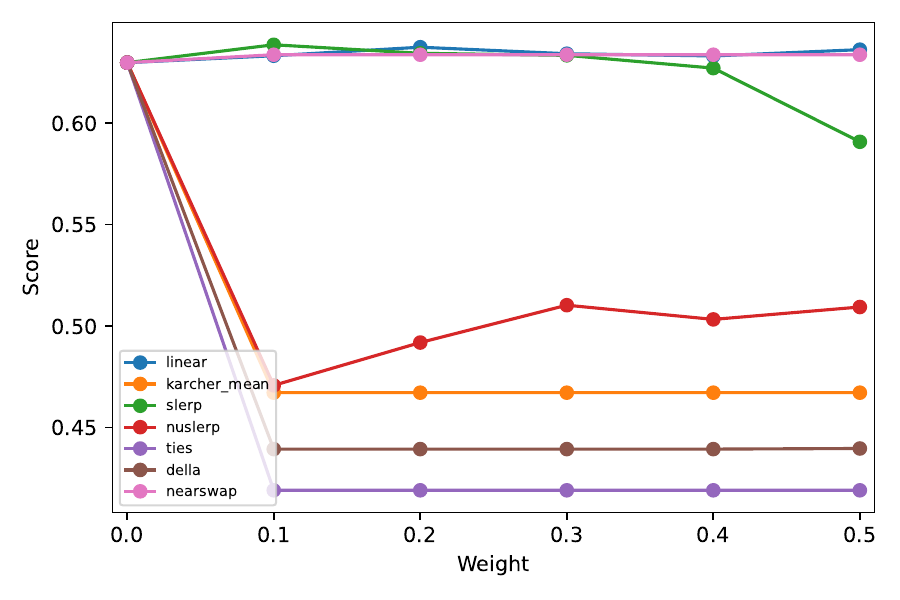}
    \caption{QWEN2-1.5B}
  \end{subfigure}
  \vspace{1em}
  \begin{subfigure}[b]{0.4\textwidth}
    \centering
    \includegraphics[width=\linewidth]{img/kekka/Qwen_Qwen2-7B/00_results_sglue_QWEN.pdf}
    \caption{QWEN2-7B}
  \end{subfigure}
  \caption{The SuperGLUE evaluation results in QWEN models.}
  \label{fig:app-sglue-qwen}
\end{figure*}

\begin{figure*}[htbp]
  \centering
  % 1行目
  \begin{subfigure}[b]{0.45\textwidth}
    \centering
    \includegraphics[width=\linewidth]{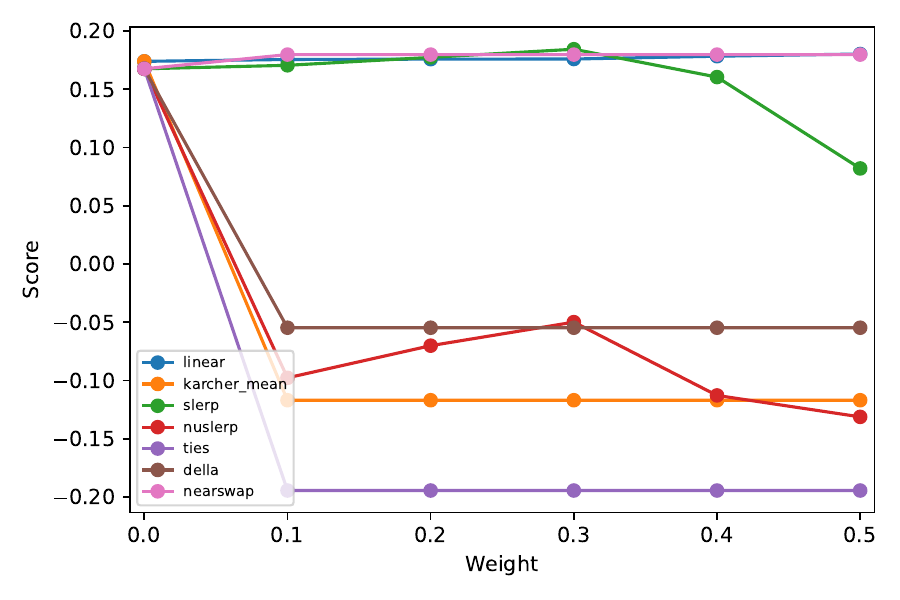}
    \caption{GPT2-small}
  \end{subfigure}
  \hfill
  \begin{subfigure}[b]{0.45\textwidth}
    \centering
    \includegraphics[width=\linewidth]{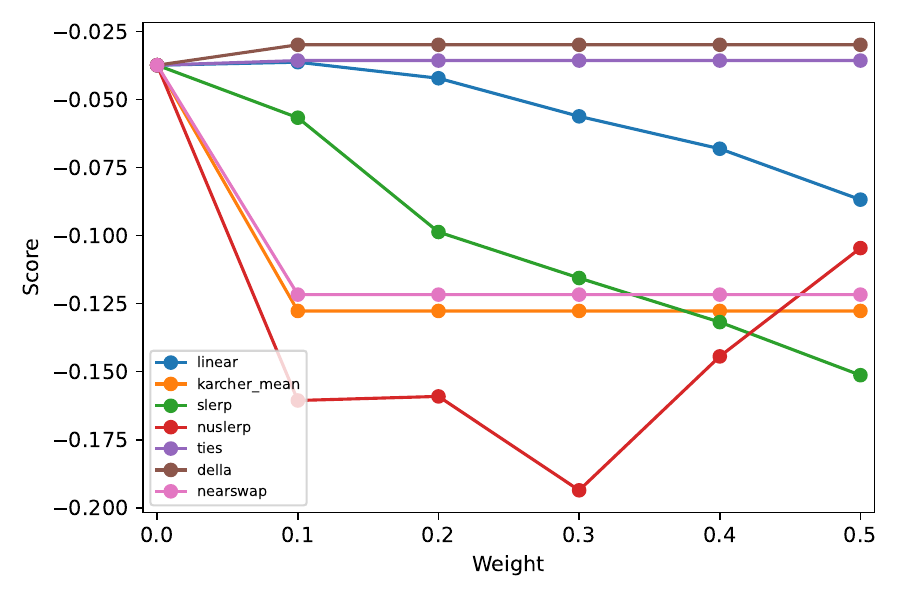}
    \caption{GPT2-medium}
  \end{subfigure}

  \vspace{1em}
  % 2行目
  \begin{subfigure}[b]{0.45\textwidth}
    \centering
    \includegraphics[width=\linewidth]{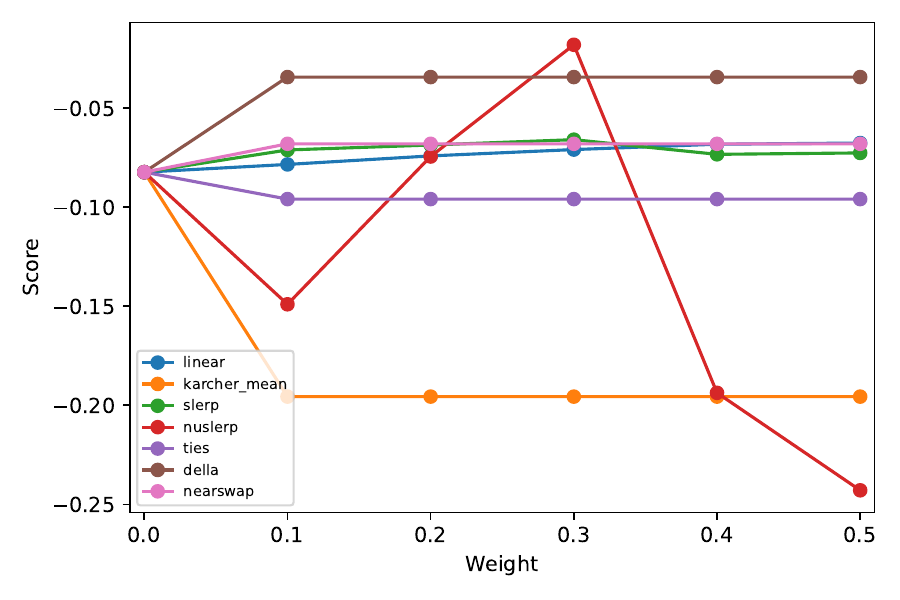}
    \caption{GPT2-large}
  \end{subfigure}
  \hfill
  \begin{subfigure}[b]{0.45\textwidth}
    \centering
    \includegraphics[width=\linewidth]{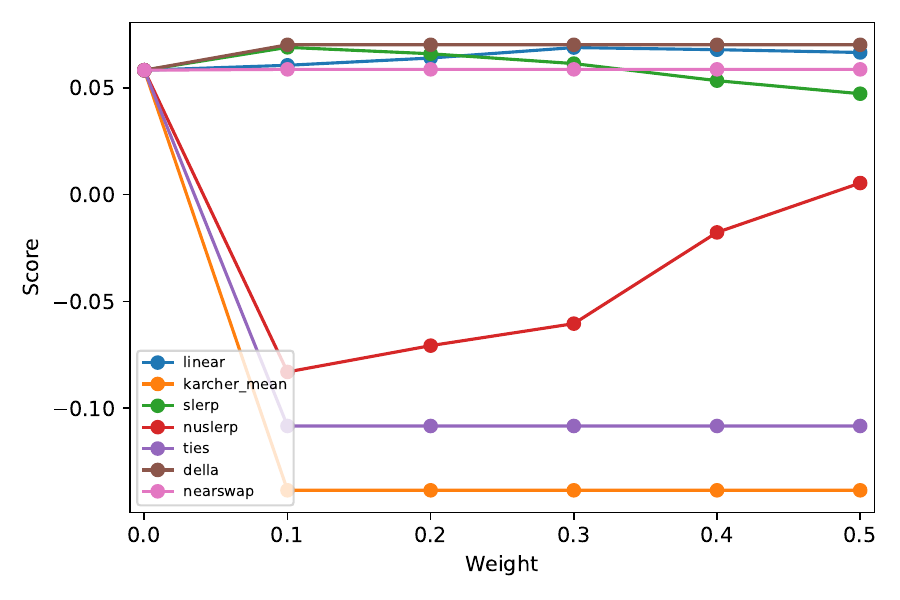}
    \caption{GPT2-XL}
  \end{subfigure}
  
  \vspace{1em}
  % 3行目
  \begin{subfigure}[b]{0.45\textwidth}
    \centering
    \includegraphics[width=\linewidth]{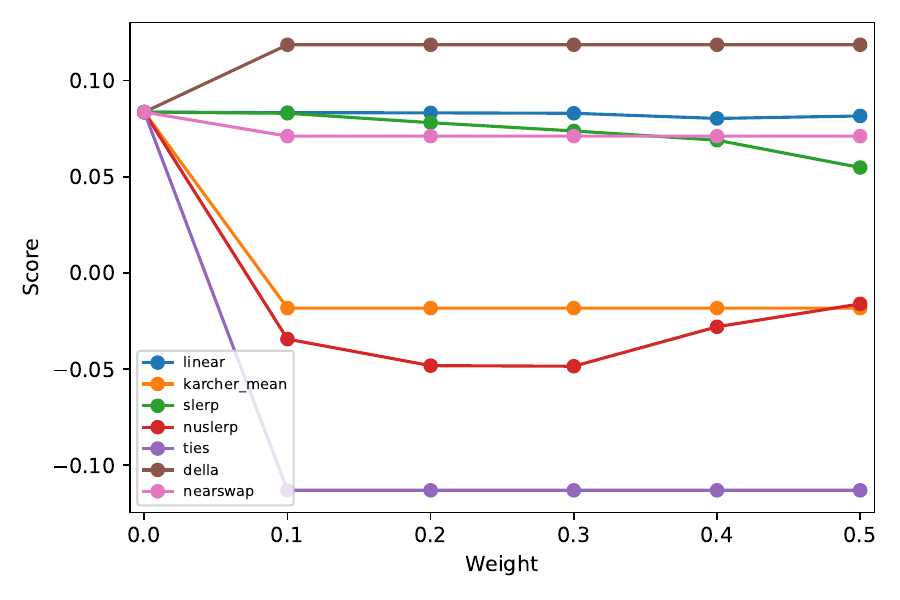}
    \caption{GPT-Neo-2.7B}
  \end{subfigure}

  \caption{The BBQ evaluation results in GPT models.}
  \label{fig:app-bbq-gpt}
\end{figure*}

\begin{figure*}[htbp]
  \centering
  % 1行目
  \begin{subfigure}[b]{0.45\textwidth}
    \centering
    \includegraphics[width=\linewidth]{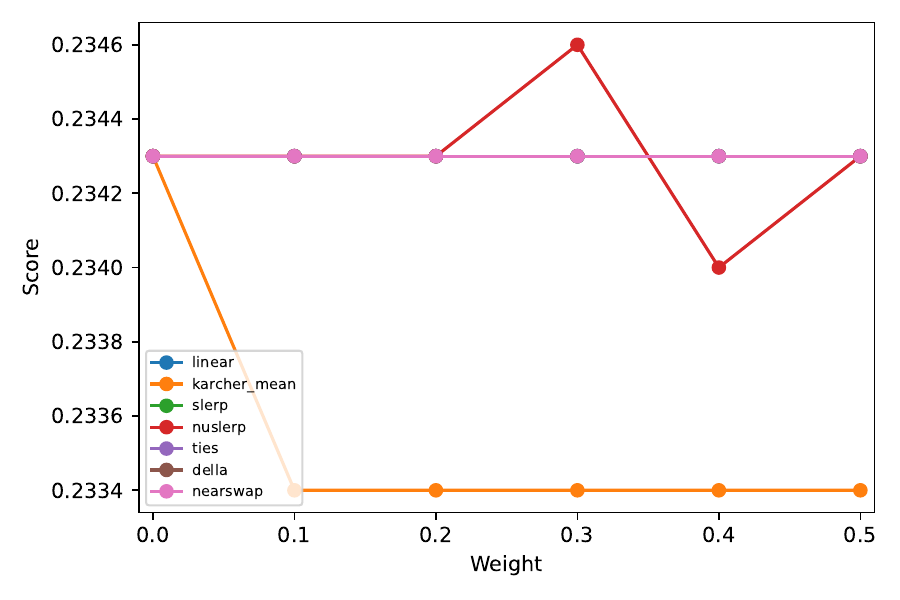}
    \caption{LLAMA2-7B}
  \end{subfigure}
  \hfill
  \begin{subfigure}[b]{0.45\textwidth}
    \centering
    \includegraphics[width=\linewidth]{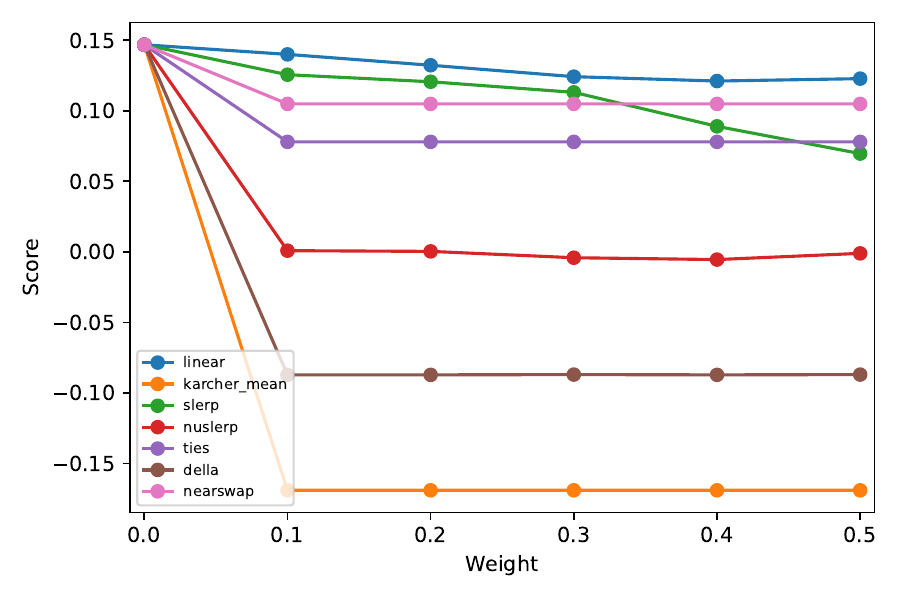}
    \caption{LLAMA-3.1-8B}
  \end{subfigure}

  \vspace{1em}
  % 2行目
  \begin{subfigure}[b]{0.45\textwidth}
    \centering
    \includegraphics[width=\linewidth]{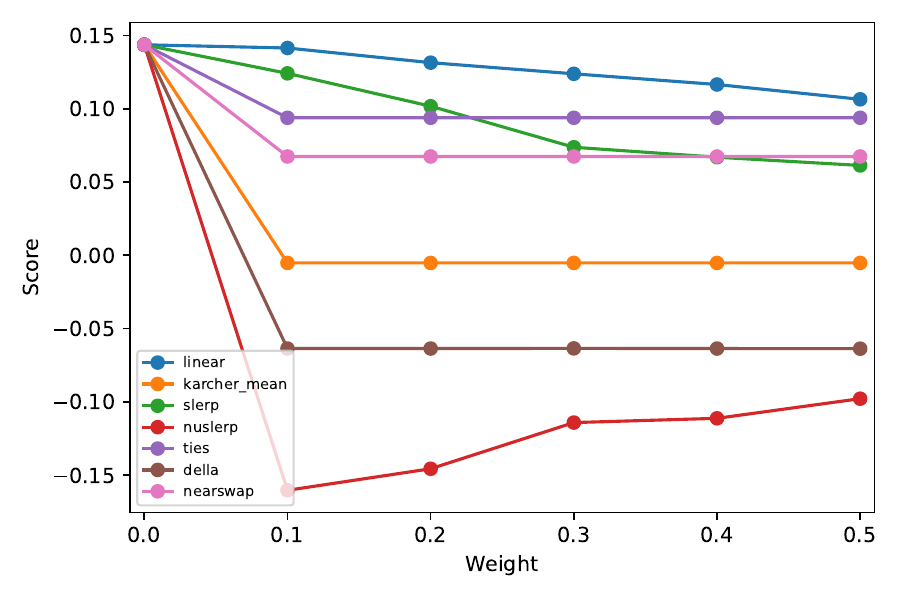}
    \caption{LLAMA-3.2-1B}
  \end{subfigure}
  \hfill
  \begin{subfigure}[b]{0.45\textwidth}
    \centering
    \includegraphics[width=\linewidth]{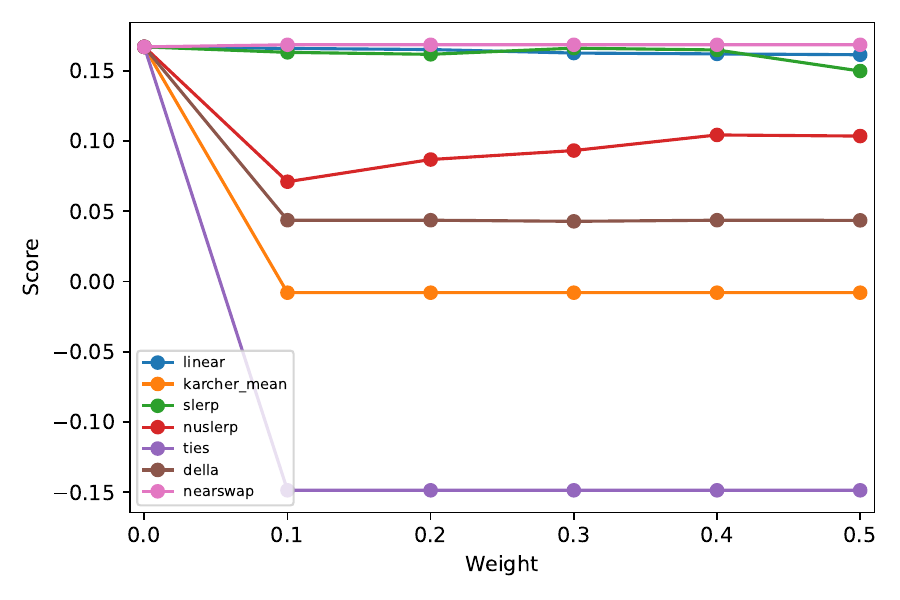}
    \caption{LLAMA-3.2-3B}
  \end{subfigure}
  \caption{The BBQ evaluation results in LLAMA models.}
  \label{fig:app-bbq-llama}
\end{figure*}

\begin{figure*}[htbp]
  \centering
  \vspace{1em}
  % 3行目
  \begin{subfigure}[b]{0.4\textwidth}
    \centering
    \includegraphics[width=\linewidth]{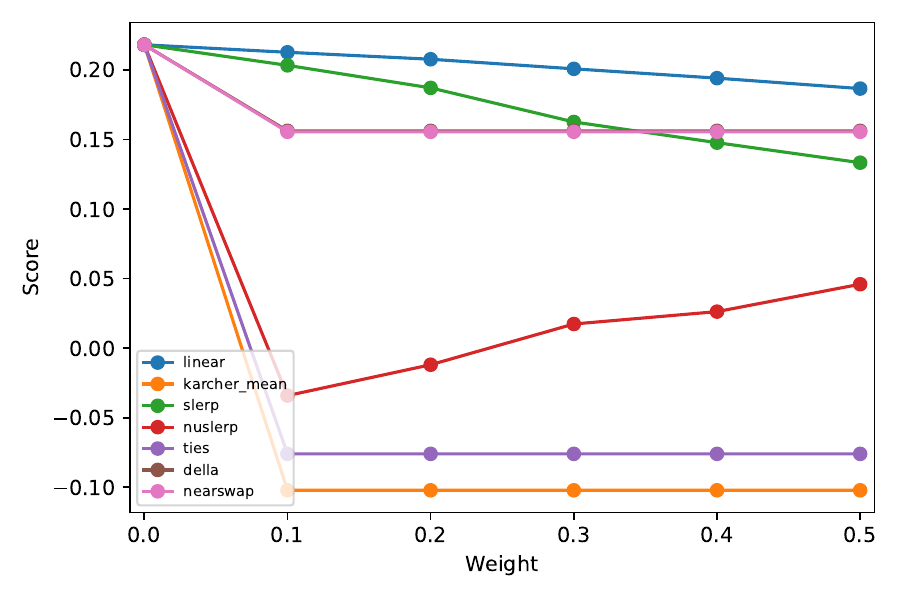}
    \caption{QWEN2-0.5B}
  \end{subfigure}
  \hfill
  \begin{subfigure}[b]{0.4\textwidth}
    \centering
    \includegraphics[width=\linewidth]{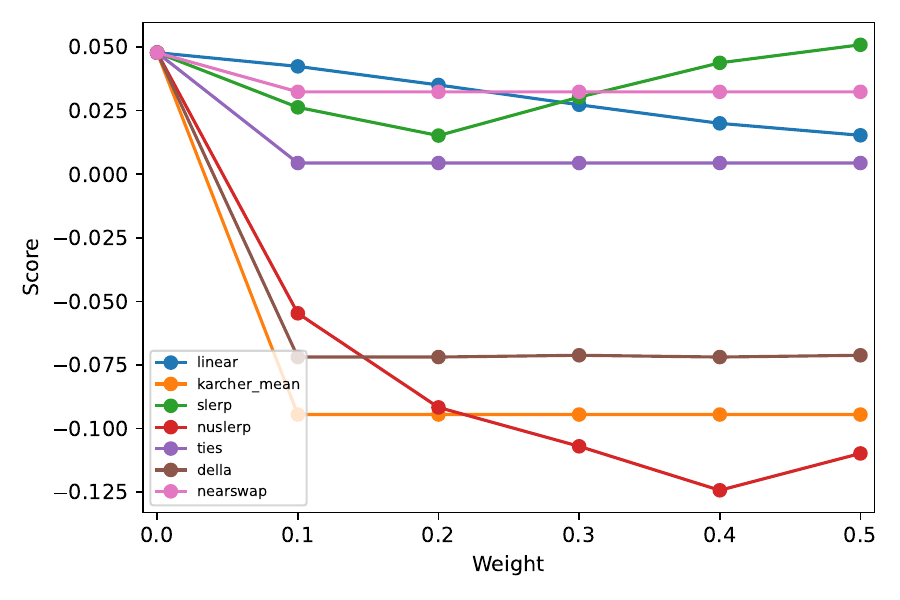}
    \caption{QWEN2-1.5B}
  \end{subfigure}
  \vspace{1em}
  \begin{subfigure}[b]{0.4\textwidth}
    \centering
    \includegraphics[width=\linewidth]{img/kekka/Qwen_Qwen2-7B/00_results_bbq_bias_QWEN.pdf}
    \caption{QWEN2-7B}
  \end{subfigure}
  \caption{The BBQ evaluation results in QWEN models.}
  \label{fig:app-bbq-qwen}
\end{figure*}

\begin{figure*}[htbp]
  \centering
  % 1行目
  \begin{subfigure}[b]{0.45\textwidth}
    \centering
    \includegraphics[width=\linewidth]{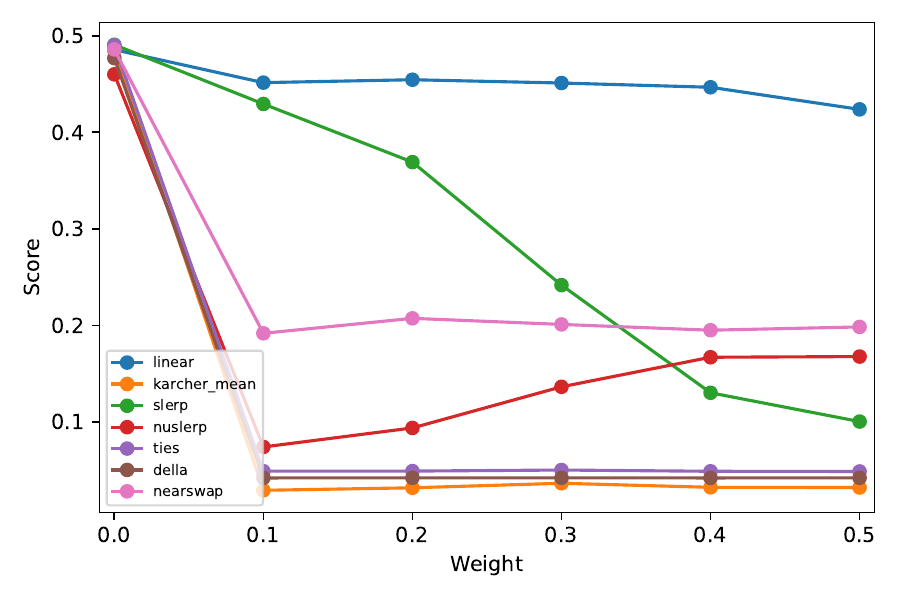}
    \caption{GPT2-small}
  \end{subfigure}
  \hfill
  \begin{subfigure}[b]{0.45\textwidth}
    \centering
    \includegraphics[width=\linewidth]{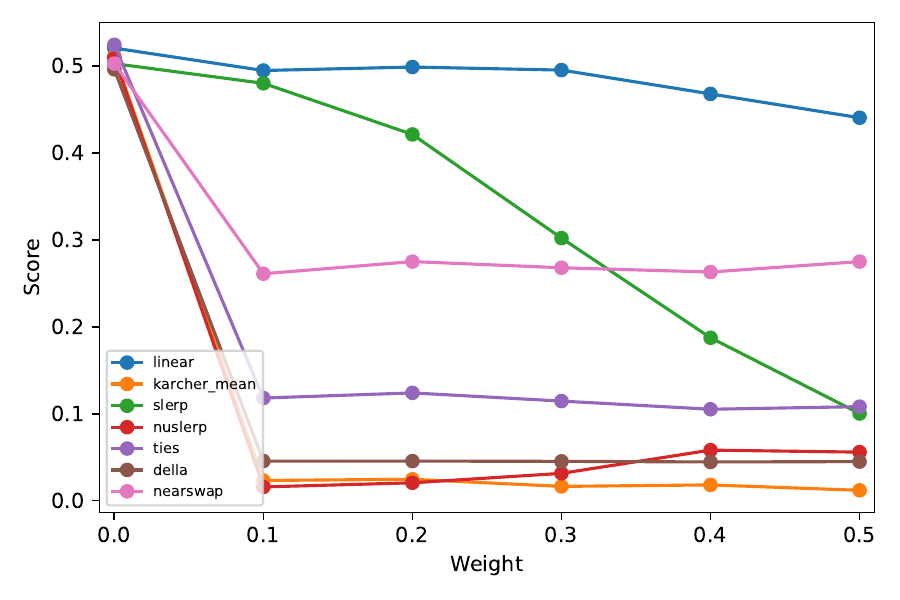}
    \caption{GPT2-medium}
  \end{subfigure}

  \vspace{1em}
  % 2行目
  \begin{subfigure}[b]{0.45\textwidth}
    \centering
    \includegraphics[width=\linewidth]{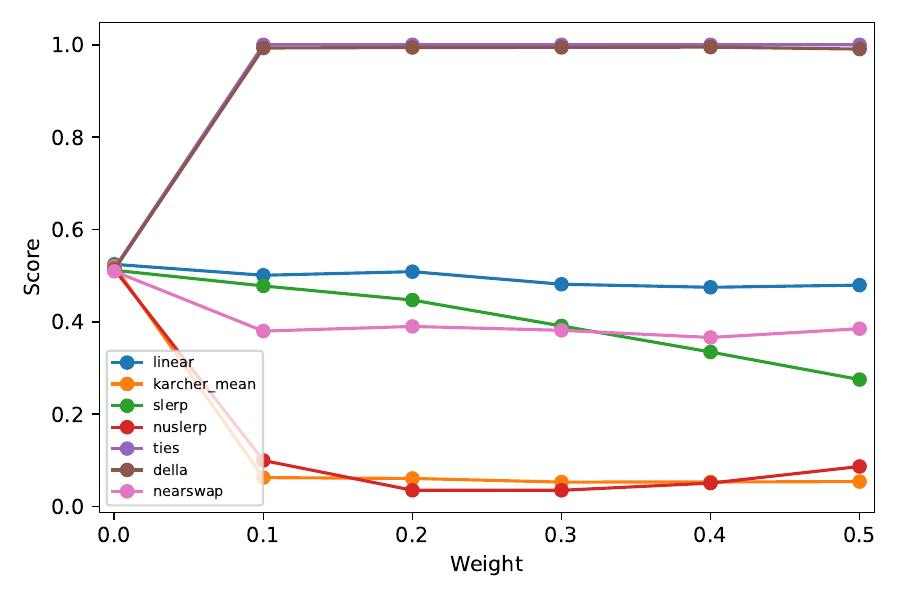}
    \caption{GPT2-large}
  \end{subfigure}
  \hfill
  \begin{subfigure}[b]{0.45\textwidth}
    \centering
    \includegraphics[width=\linewidth]{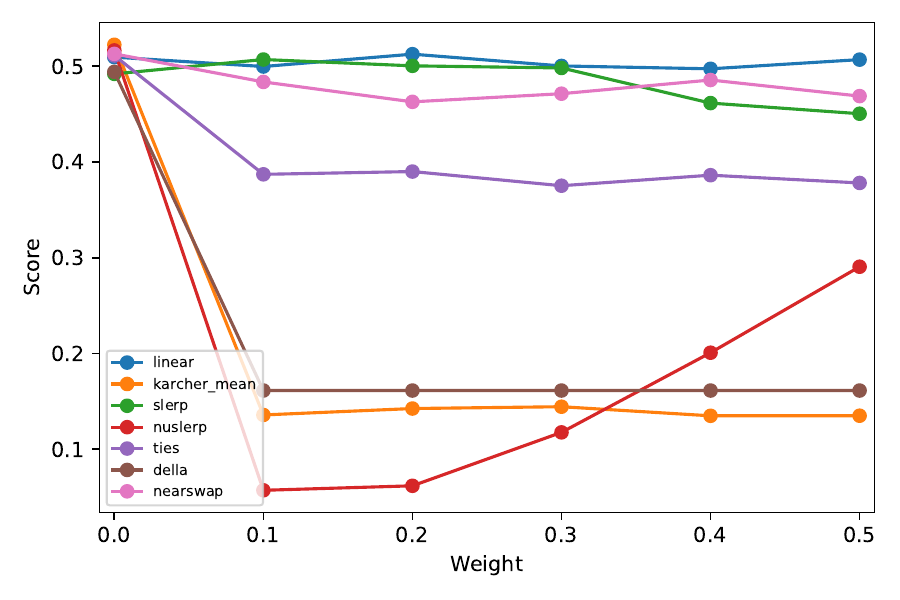}
    \caption{GPT2-XL}
  \end{subfigure}
  
  \vspace{1em}
  % 3行目
  \begin{subfigure}[b]{0.45\textwidth}
    \centering
    \includegraphics[width=\linewidth]{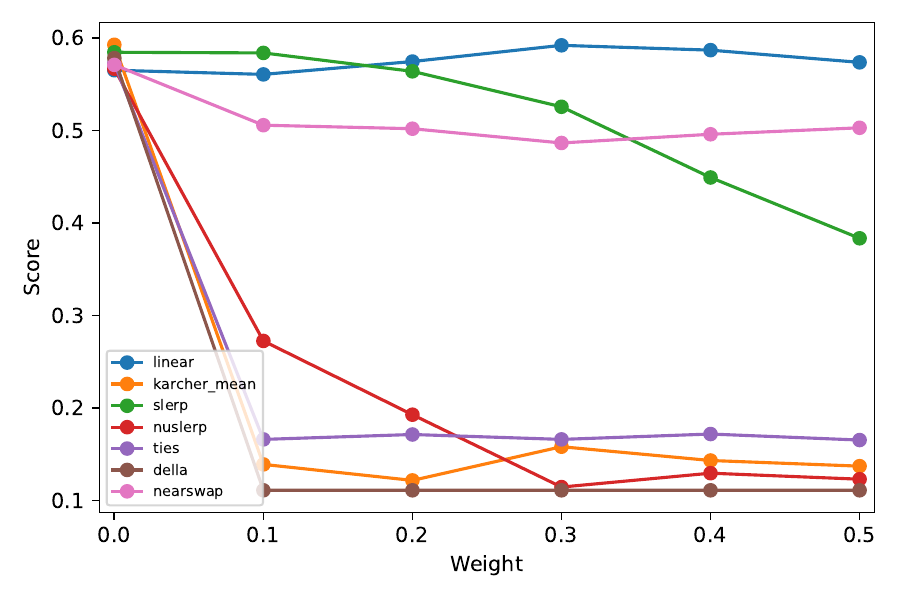}
    \caption{GPT-Neo-2.7B}
  \end{subfigure}

  \caption{The BOLD evaluation results in GPT models.}
  \label{fig:app-bold-gpt}
\end{figure*}

\begin{figure*}[htbp]
  \centering
  % 1行目
  \begin{subfigure}[b]{0.45\textwidth}
    \centering
    \includegraphics[width=\linewidth]{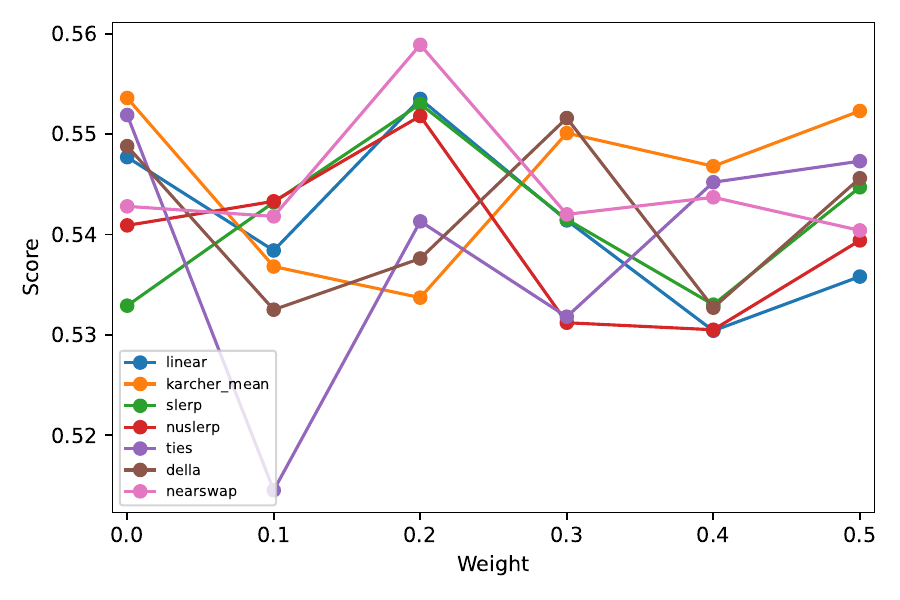}
    \caption{LLAMA2-7B}
  \end{subfigure}
  \hfill
  \begin{subfigure}[b]{0.45\textwidth}
    \centering
    \includegraphics[width=\linewidth]{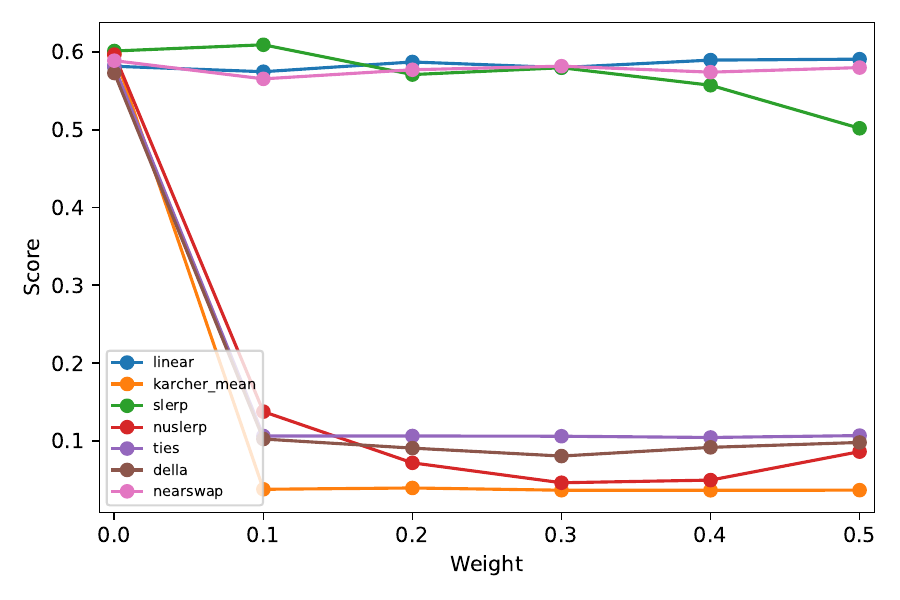}
    \caption{LLAMA-3.1-8B}
  \end{subfigure}

  \vspace{1em}
  % 2行目
  \begin{subfigure}[b]{0.45\textwidth}
    \centering
    \includegraphics[width=\linewidth]{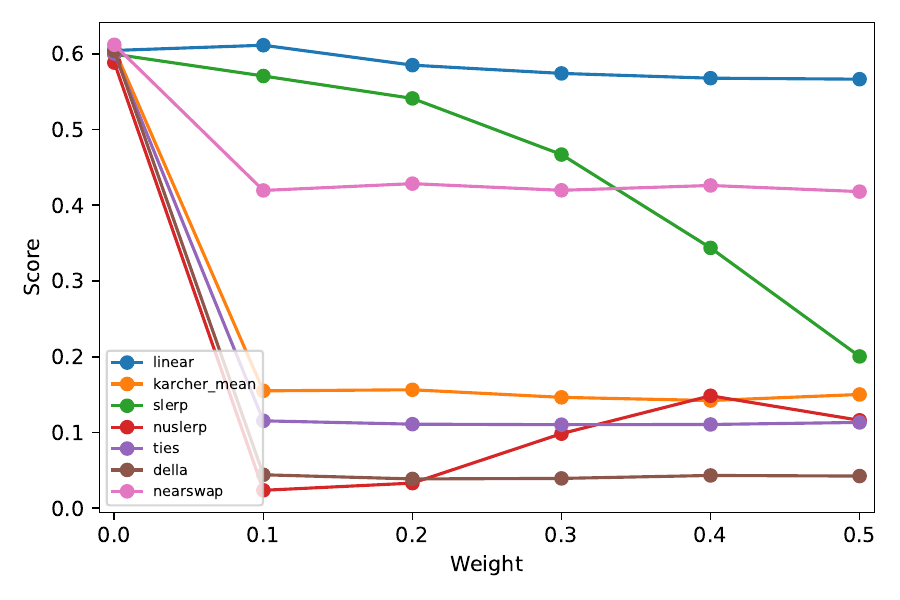}
    \caption{LLAMA-3.2-1B}
  \end{subfigure}
  \hfill
  \begin{subfigure}[b]{0.45\textwidth}
    \centering
    \includegraphics[width=\linewidth]{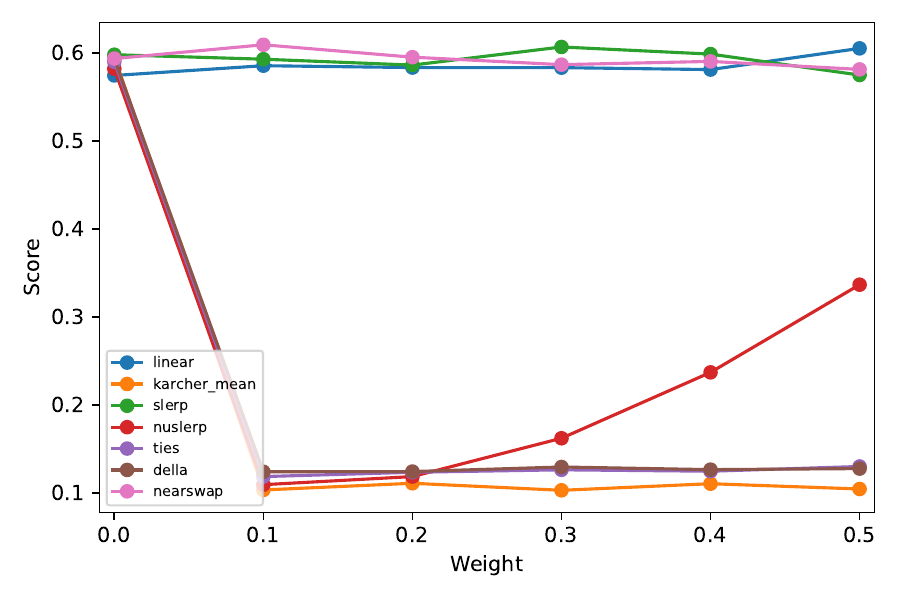}
    \caption{LLAMA-3.2-3B}
  \end{subfigure}
  \caption{The BOLD evaluation results in LLAMA models.}
  \label{fig:app-bold-llama}
\end{figure*}

\begin{figure*}[htbp]
  \centering
  \vspace{1em}
  % 3行目
  \begin{subfigure}[b]{0.4\textwidth}
    \centering
    \includegraphics[width=\linewidth]{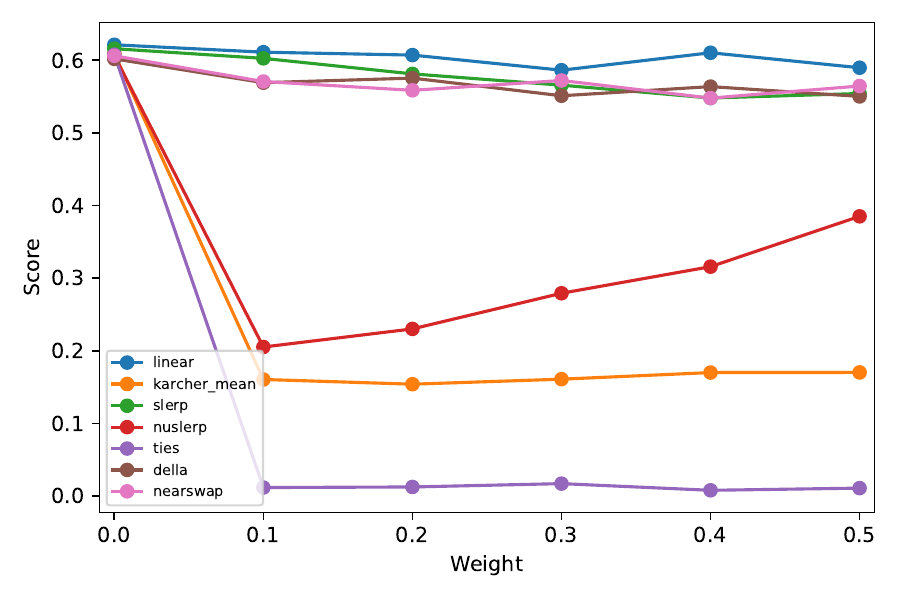}
    \caption{QWEN2-0.5B}
  \end{subfigure}
  \hfill
  \begin{subfigure}[b]{0.4\textwidth}
    \centering
    \includegraphics[width=\linewidth]{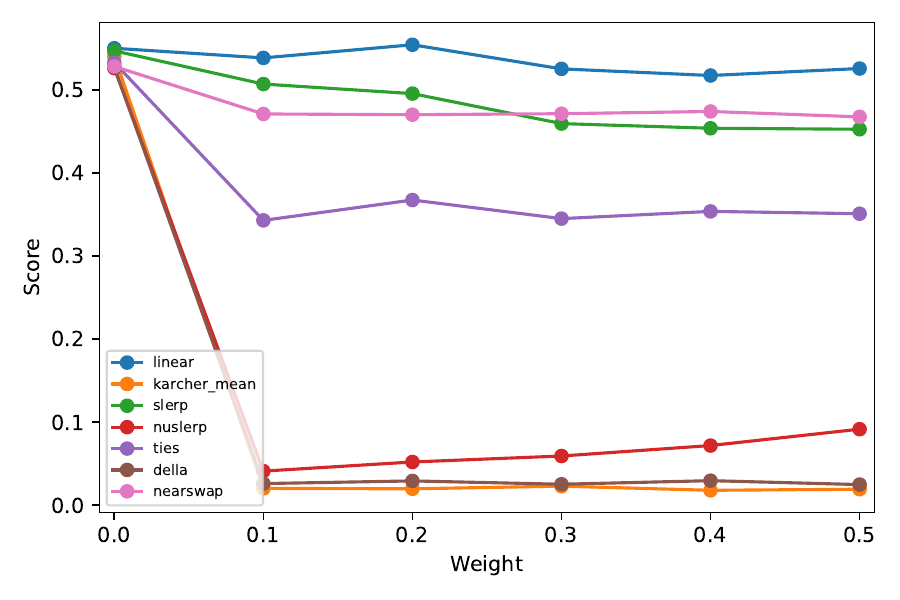}
    \caption{QWEN2-1.5B}
  \end{subfigure}
  \vspace{1em}
  \begin{subfigure}[b]{0.4\textwidth}
    \centering
    \includegraphics[width=\linewidth]{img/kekka/Qwen_Qwen2-7B/00_results_bold_QWEN.pdf}
    \caption{QWEN2-7B}
  \end{subfigure}
  \caption{The BOLD evaluation results in QWEN models.}
  \label{fig:app-bold-qwen}
\end{figure*}

\begin{figure*}[htbp]
  \centering
  % 1行目
  \begin{subfigure}[b]{0.45\textwidth}
    \centering
    \includegraphics[width=\linewidth]{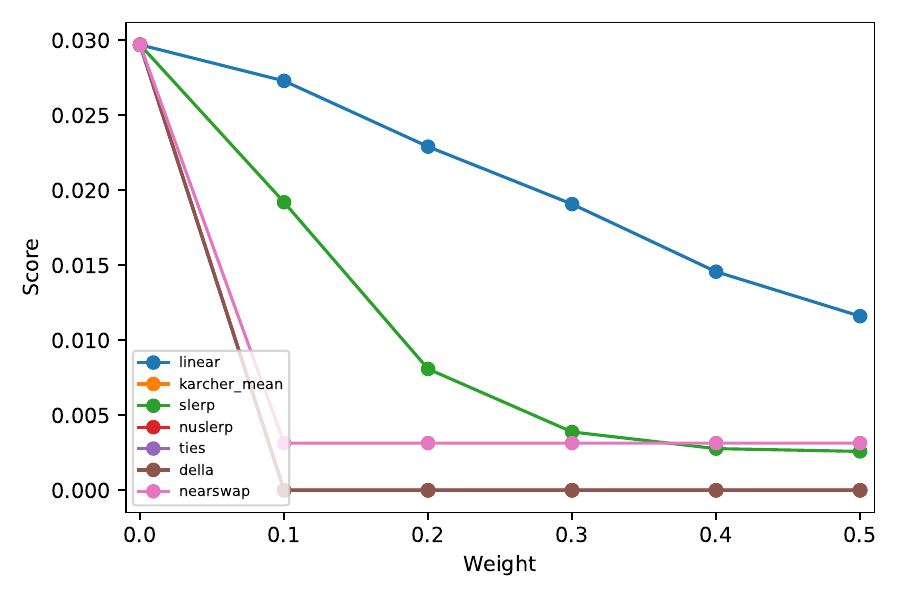}
    \caption{GPT2-small}
  \end{subfigure}
  \hfill
  \begin{subfigure}[b]{0.45\textwidth}
    \centering
    \includegraphics[width=\linewidth]{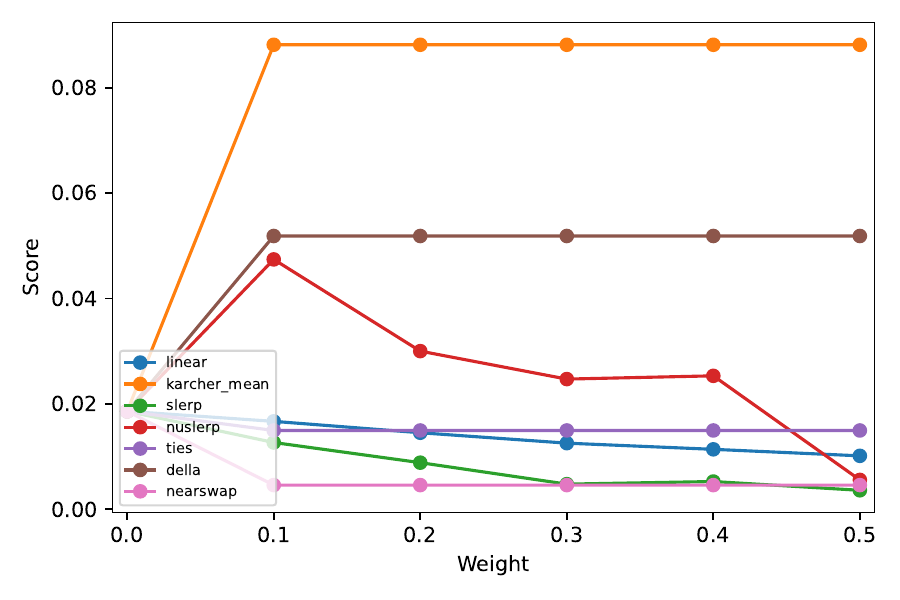}
    \caption{GPT2-medium}
  \end{subfigure}

  \vspace{1em}
  % 2行目
  \begin{subfigure}[b]{0.45\textwidth}
    \centering
    \includegraphics[width=\linewidth]{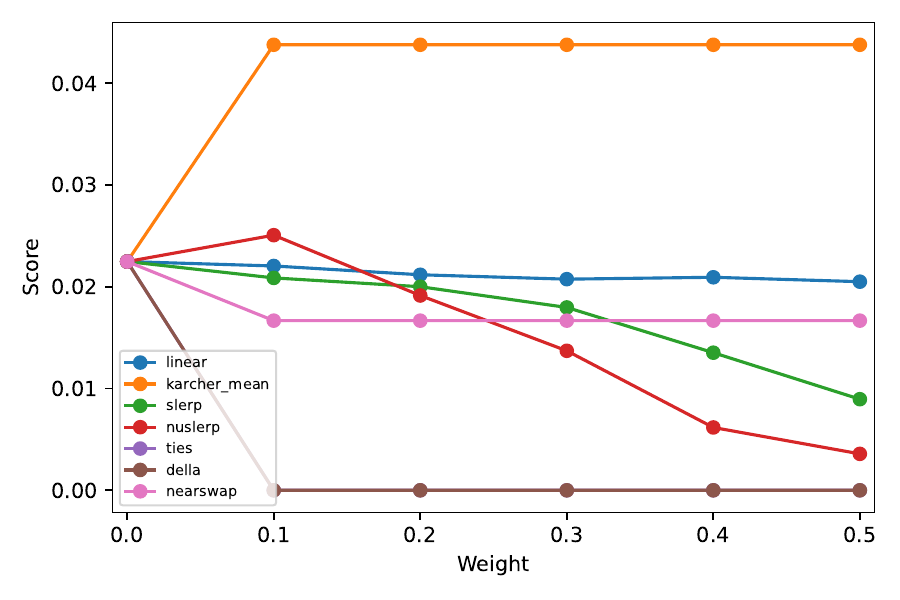}
    \caption{GPT2-large}
  \end{subfigure}
  \hfill
  \begin{subfigure}[b]{0.45\textwidth}
    \centering
    \includegraphics[width=\linewidth]{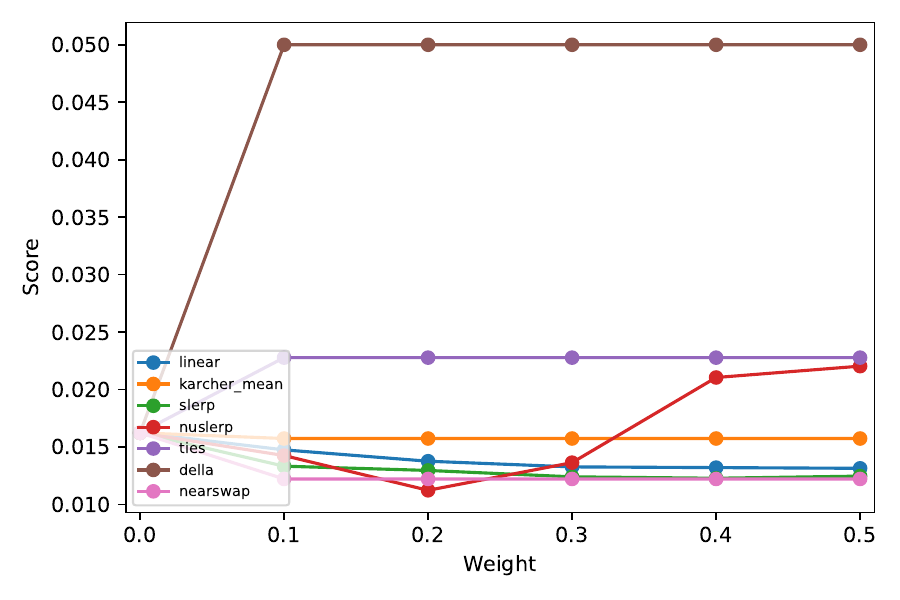}
    \caption{GPT2-XL}
  \end{subfigure}
  
  \vspace{1em}
  % 3行目
  \begin{subfigure}[b]{0.45\textwidth}
    \centering
    \includegraphics[width=\linewidth]{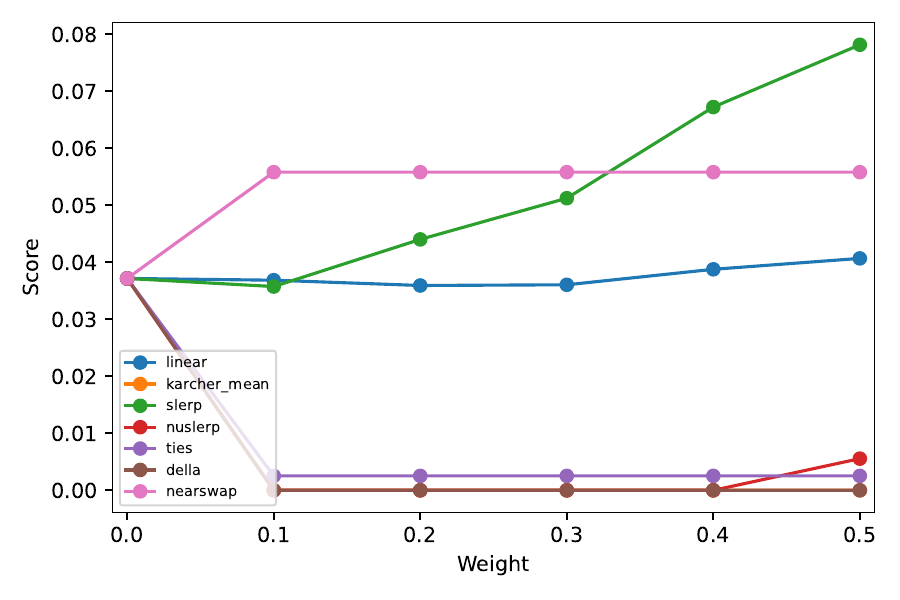}
    \caption{GPT-Neo-2.7B}
  \end{subfigure}

  \caption{The HONEST evaluation results in GPT models.}
  \label{fig:app-honest-gpt}
\end{figure*}

\begin{figure*}[htbp]
  \centering
  % 1行目
  \begin{subfigure}[b]{0.45\textwidth}
    \centering
    \includegraphics[width=\linewidth]{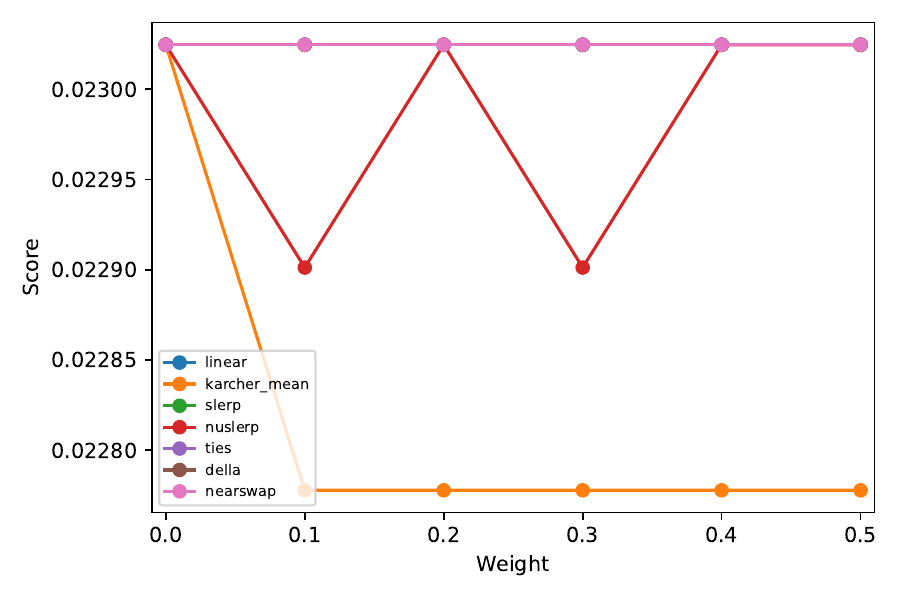}
    \caption{LLAMA2-7B}
  \end{subfigure}
  \hfill
  \begin{subfigure}[b]{0.45\textwidth}
    \centering
    \includegraphics[width=\linewidth]{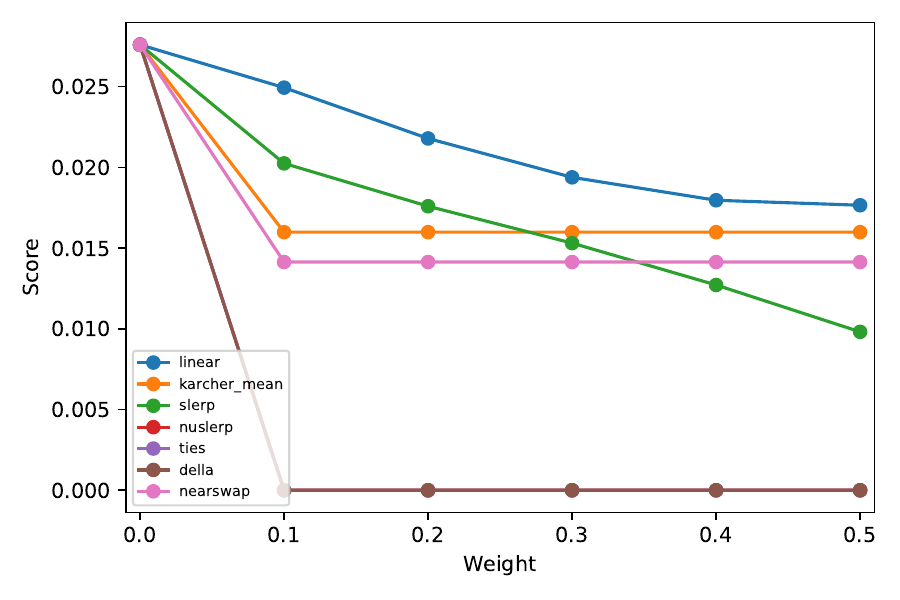}
    \caption{LLAMA-3.1-8B}
  \end{subfigure}

  \vspace{1em}
  % 2行目
  \begin{subfigure}[b]{0.45\textwidth}
    \centering
    \includegraphics[width=\linewidth]{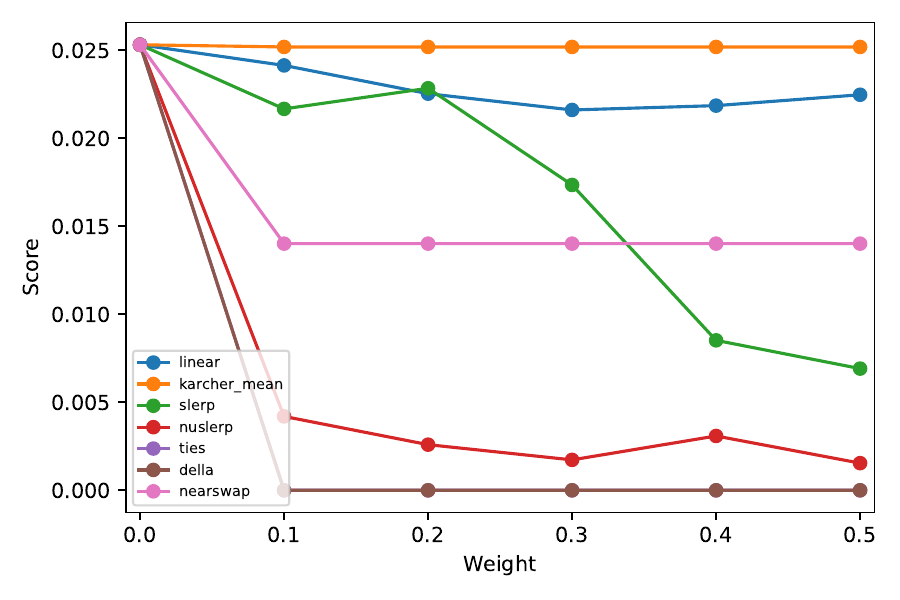}
    \caption{LLAMA-3.2-1B}
  \end{subfigure}
  \hfill
  \begin{subfigure}[b]{0.45\textwidth}
    \centering
    \includegraphics[width=\linewidth]{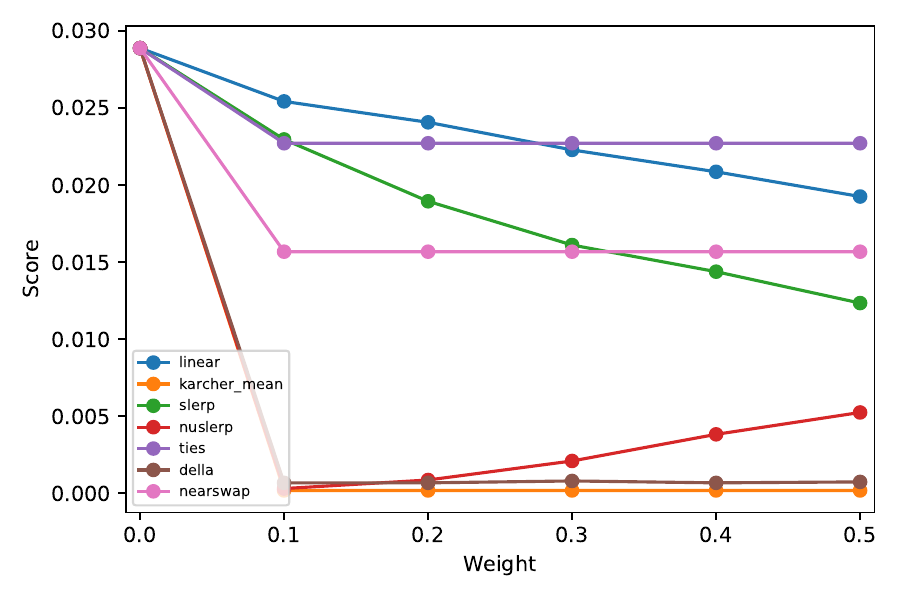}
    \caption{LLAMA-3.2-3B}
  \end{subfigure}
  \caption{The HONEST evaluation results in LLAMA models.}
  \label{fig:app-honest-llama}
\end{figure*}

\begin{figure*}[htbp]
  \centering
  \vspace{1em}
  % 3行目
  \begin{subfigure}[b]{0.4\textwidth}
    \centering
    \includegraphics[width=\linewidth]{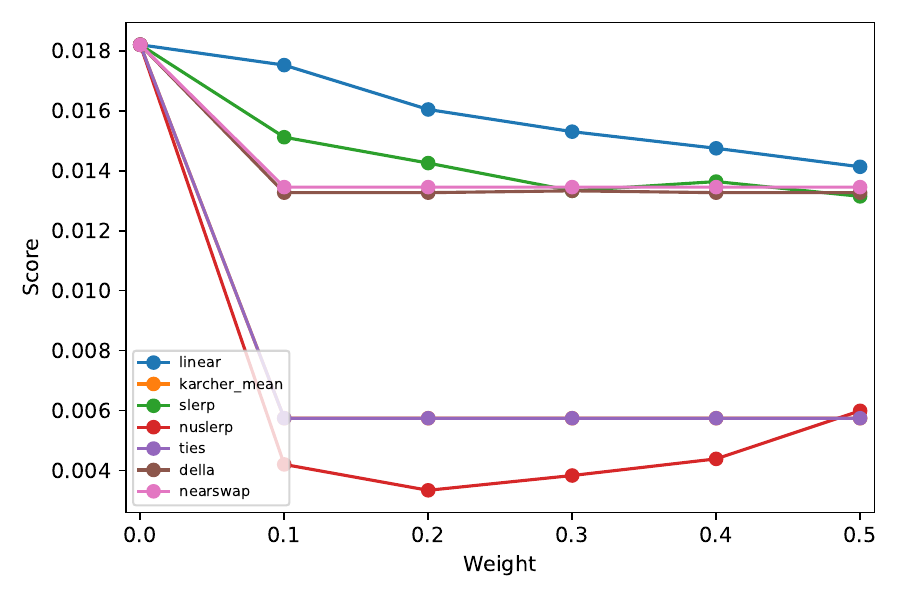}
    \caption{QWEN2-0.5B}
  \end{subfigure}
  \hfill
  \begin{subfigure}[b]{0.4\textwidth}
    \centering
    \includegraphics[width=\linewidth]{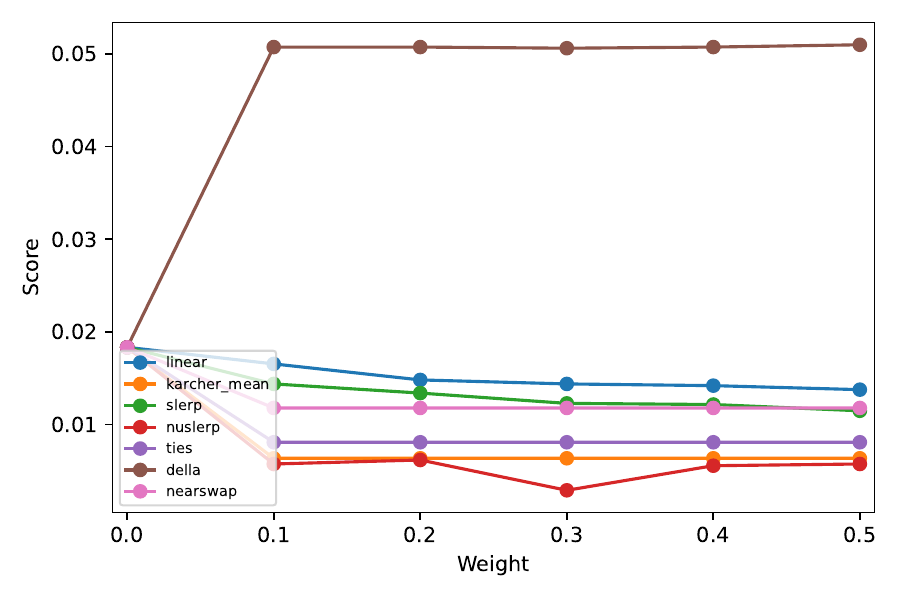}
    \caption{QWEN2-1.5B}
  \end{subfigure}
  \vspace{1em}
  \begin{subfigure}[b]{0.4\textwidth}
    \centering
    \includegraphics[width=\linewidth]{img/kekka/Qwen_Qwen2-7B/00_results_honest_QWEN.pdf}
    \caption{QWEN2-7B}
  \end{subfigure}
  \caption{The HONEST evaluation results in QWEN models.}
  \label{fig:app-honest-qwen}
\end{figure*}

This section shows the results of each LLM
evaluated with SuperGLUE, BBQ, BOLD, and HONEST benchmarks.
The results are shown in
Figure \ref{fig:app-sglue-gpt} (GPT on SuperGLUE), \ref{fig:app-sglue-llama} (LLAMA on SuperGLUE), \ref{fig:app-sglue-qwen} (Qwen on SuperGLUE),
\ref{fig:app-bbq-gpt} (GPT on BBQ), \ref{fig:app-bbq-llama} (LLAMA on BBQ), \ref{fig:app-bbq-qwen} (Qwen on BBQ),
\ref{fig:app-bold-gpt} (GPT on BOLD), \ref{fig:app-bold-llama} (LLAMA on BOLD), \ref{fig:app-bold-qwen} (Qwen on BOLD),
\ref{fig:app-honest-gpt} (GPT on HONEST), \ref{fig:app-honest-llama} (LLAMA on HONEST), and \ref{fig:app-honest-qwen} (Qwen on HONEST).

\end{document}